\documentclass{article}

\PassOptionsToPackage{numbers,sort&compress}{natbib}
\usepackage[final,main]{neurips_2026}

\usepackage[utf8]{inputenc}
\usepackage[T1]{fontenc}
\usepackage[hidelinks]{hyperref}
\usepackage{url}
\usepackage{booktabs}
\usepackage{amsfonts}
\usepackage{amsmath}
\usepackage{amssymb}
\usepackage{nicefrac}
\usepackage{microtype}
\usepackage{xcolor}
\usepackage{graphicx}
\usepackage{subcaption}
\usepackage{xspace}
\usepackage[capitalize,noabbrev]{cleveref}
\usepackage{algorithm}  
\usepackage{algpseudocode}
\usepackage{enumitem}
\usepackage{wrapfig}
\usepackage{float}
\usepackage{caption}
\newcommand{\ours}{WMMs\xspace}
\newcommand{\ourssing}{WMM\xspace}

\newcommand{\ourslong}{World Motion Models\xspace}

\title{\ourslong: \\Flexible Sequence Modeling of SE(3) Trajectories}

\author{
  \normalfont
  \begin{tabular*}{\dimexpr\textwidth-2\tabcolsep\relax}{@{\extracolsep{\fill}}cccc@{}}
    \textbf{Jiahui Lei} & \textbf{Qianqian Wang} & \textbf{Trevor Darrell} & \textbf{Angjoo Kanazawa} \\[6pt]
    UC Berkeley & Harvard University & UC Berkeley & UC Berkeley
  \end{tabular*} \\[1.2em]
  \texttt{\{leijh,trevordarrell,kanazawa\}@berkeley.edu}\hspace{6pt}\texttt{qwang@seas.harvard.edu}
}

\begin{document}

\maketitle

\vspace{-2em}
\begin{abstract}

Equipping artificial agents with spatial intelligence requires a comprehensive generative prior over the dynamic 3D world.
We propose \ourslong (\ours) that capture \textbf{``what was, is, and will be where across time''} via sparse \emph{SE(3) pose trajectories}.
\ours are built on the observation that elements of dynamic scenes can be well approximated by a set of rigid SE(3) trajectories, a minimal yet expressive primitive for 4D modeling. 
This representation unifies articulated objects, human bodies, hand-object interactions, piecewise-rigid scene dynamics, camera motion, and even robot states and actions into a single shared space. Given this representation, we cast the joint distribution of these entities as a flexible sequence modeling problem, utilizing flow-matching with per-token noise levels. Coupled with a context token mechanism for non-sequential conditioning, this formulation supports any-to-any marginal conditioning across an arbitrary number of entities and time steps. Tasks such as future prediction, motion infilling, model-predictive control, inverse kinematics, cross-embodiment retargeting, and policy learning all reduce to the application of different masks over the same network. Experiments on 6 diverse applications of 3D vision and robotics demonstrate the versatility and flexibility of \ours with strong performance.

\end{abstract}

\begin{center}
    \small\url{https://jiahuilei.com/projects/wmm}
\end{center}

\begin{figure}[H]
    \centering
    \includegraphics[width=\linewidth]{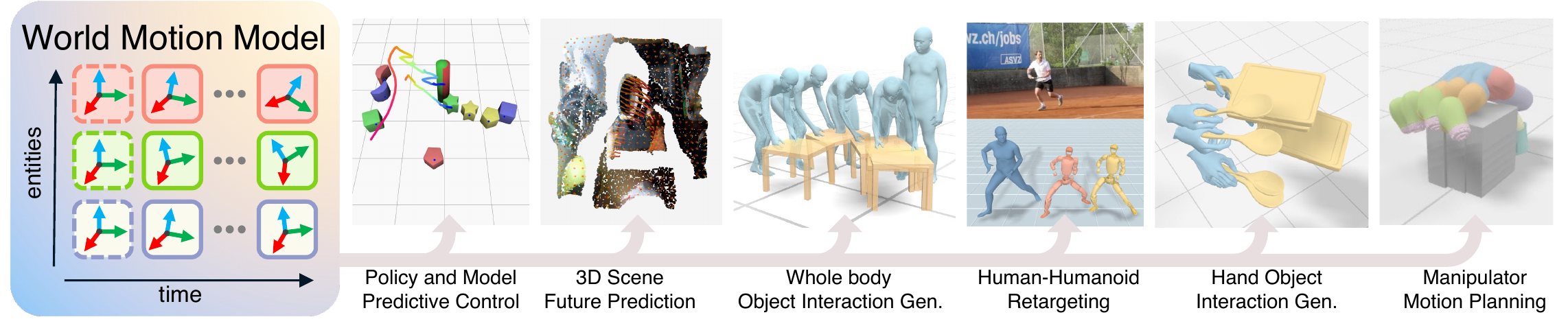}
    \caption{\textbf{\ourslong} represent the 4D world across diverse entities and prediction modes in a unified manner, and can be flexibly conditioned for various downstream tasks.}
    \label{fig:teaser}
\end{figure}
\vspace{-1em}

\section{Introduction}
\label{sec:intro}
The world that surrounds us is inherently three-dimensional and highly dynamic—a 4D space where perception and interaction are fundamentally tied to movement. Physical actions are achieved by moving limbs or end-effectors, generating forces that translate into the motion of objects, while other agents dynamically navigate and act within this same environment. As humans, we naturally develop a strong prior over these moving entities: we intuitively anticipate how others will move, plan the physical motions required to achieve a desired goal, and foresee the physical consequences of our own actions.
To act and reason in the physical world, intelligent systems benefit from the same capability: a generative prior over the 4D world that captures the evolution of ``what'' and ``where'' across time.

In this work, we present a unified generative model which provides a general prior over the dynamic 3D world. Our approach is highly flexible with minimal assumptions and handles unstructured scenes and arbitrary object categories. By supporting various prompting modes, a single model enables diverse inference tasks ranging from world modeling (e.g., action-conditioned/free future prediction) to complex planning (e.g., motion infilling, inverse dynamics).

To build such a prior, we first need a unified representation that expresses the dynamic world in a physically grounded manner.
Our insight is that complex, dense motion of the physical world can be compactly approximated by a sparse set of moving coordinate frames --- specifically $SE(3)$ trajectories. Most entities of interest naturally reduce to this format: the dense flow of thousands of mesh vertices on a human or robot collapses into tens of joint trajectories~\cite{loper2015smpl,huang2026pointworld}; cameras and rigid objects track as single moving frames; and even non-rigid scene dynamics can be represented as interpolations of a small $SE(3)$ motion basis~\cite{wang2024som,kratimenos2024dynmf,lei2024mosca}. %
We therefore adopt the $SE(3)$ trajectory as our basic language for \emph{what is where across time}.
These $SE(3)$ trajectories subsume prior point trajectories in $\mathbb{R}^3$ but can be far sparser, replacing, for instance, the flow of thousands of mesh vertices with the tens of joint poses of a SMPL human~\cite{loper2015smpl} or a robot~\cite{huang2026pointworld}. We further impose no category-specific structure: every entity is treated equally as a pose proxy, generalizing class-bound representations (SMPL~\cite{loper2015smpl}, MANO~\cite{romero2017mano}, URDFs) to any unstructured scene.
It is also orthogonal to video-based world modeling~\cite{blattmann2023svd,yang2024cogvideox,nvidia2025cosmos,assran2025vjepa2,zhou2024dinowm}, which captures dynamics implicitly through color changes on a screen, entangling the underlying 3D motion with appearance, lighting, and viewpoint changes. In short, the $SE(3)$ trajectory provides a highly compact, physically grounded 3D abstraction for modeling arbitrary motion. %

The question now becomes: how do we learn a generative prior over these $SE(3)$ trajectories? Since motion naturally unfolds over time, this is inherently a sequence modeling problem. Traditionally, this is treated as a strictly causal task: a standard world model or dynamics model predicts future states conditioned on past history and actions. However, humans can perform non-causal inference as well, such as determining the actions required to achieve a specific goal (inverse dynamics) or planning intermediate motions between waypoints. The inference type varies widely across applications. To support these diverse inference schemes, we propose \ourslong (\ours),  flow-matching generative models with per-token noise levels. Inspired by Diffusion Forcing~\cite{chen2024diffusionforcing} and discrete diffusion~\cite{nie2025large,arriola2025blockdiffusion}, \ours treat each 6D $SE(3)$ pose at a given time step as an independent token, assign each its own noise level, and denoise the entire token field in a single shared forward pass.

Once trained, a \ourssing admits inference over arbitrary numbers of entities and time steps simply by dictating which tokens are clean (the condition) and which are noised (the prediction). Given past tokens, it acts as a future predictor $p(s'|s)$; applied to action tokens, it becomes a feed-forward policy $p(a|s)$; and when both past states and actions are provided, it acts as an action-conditioned dynamics model $p(s'|s, a)$ supporting model-predictive control. Ultimately, \ours realize any $p(\text{unknown}|\text{known})$. By simply changing what is known, the exact same network transitions seamlessly between serving as a policy, an action prior, and a world model.

While standard trajectory modeling assumes a strictly consecutive temporal horizon, many practical tasks require reasoning beyond it. For example, planning a complex interaction often relies on an abstract goal state situated arbitrarily far in the future. Similarly, tasks like cross-embodiment retargeting require inferring correspondences between trajectories, such as those of a human and a robot. Providing a few paired demonstration frames from outside the current episode serves as an in-context template that tells the model ``what is what.'' To support these non-consecutive anchors, we enhance the flexibility of \ours by introducing a set of \emph{context $SE(3)$ tokens}. Rather than being bound to the continuous physical timeline of the current episode, they are assigned to a decoupled context timeline with its own learnable positional embeddings. Mechanically, these context tokens are treated identically to real trajectory tokens, meaning they can seamlessly condition the model or be generated themselves.

Together, this general $SE(3)$ formulation and our flexible sequence modeling enable \ours to demonstrate broad versatility across applications, including robot policies and model-predictive control (MPC), scene future prediction, %
human-object interaction generation, %
and multi-embodiment humanoid retargeting %
(Sec.~\ref{sec:exp:lt}), all with SOTA performance.

\noindent \textbf{Contributions.} We propose a general formulation that uses the sparse $SE(3)$ trajectory as a unified representation for 4D world modeling. 
To learn a flexible generative prior over this space, we introduce \ours, flexible flow-matching models with per-token noise levels. By supporting any-to-any marginal conditioning alongside a context token mechanism, \ours deliver strong performance across diverse 3D vision and robotics tasks. %

\section{\ourslong}

\label{sec:method}
\begin{figure}[!b]
    \centering
    \includegraphics[width=\linewidth]{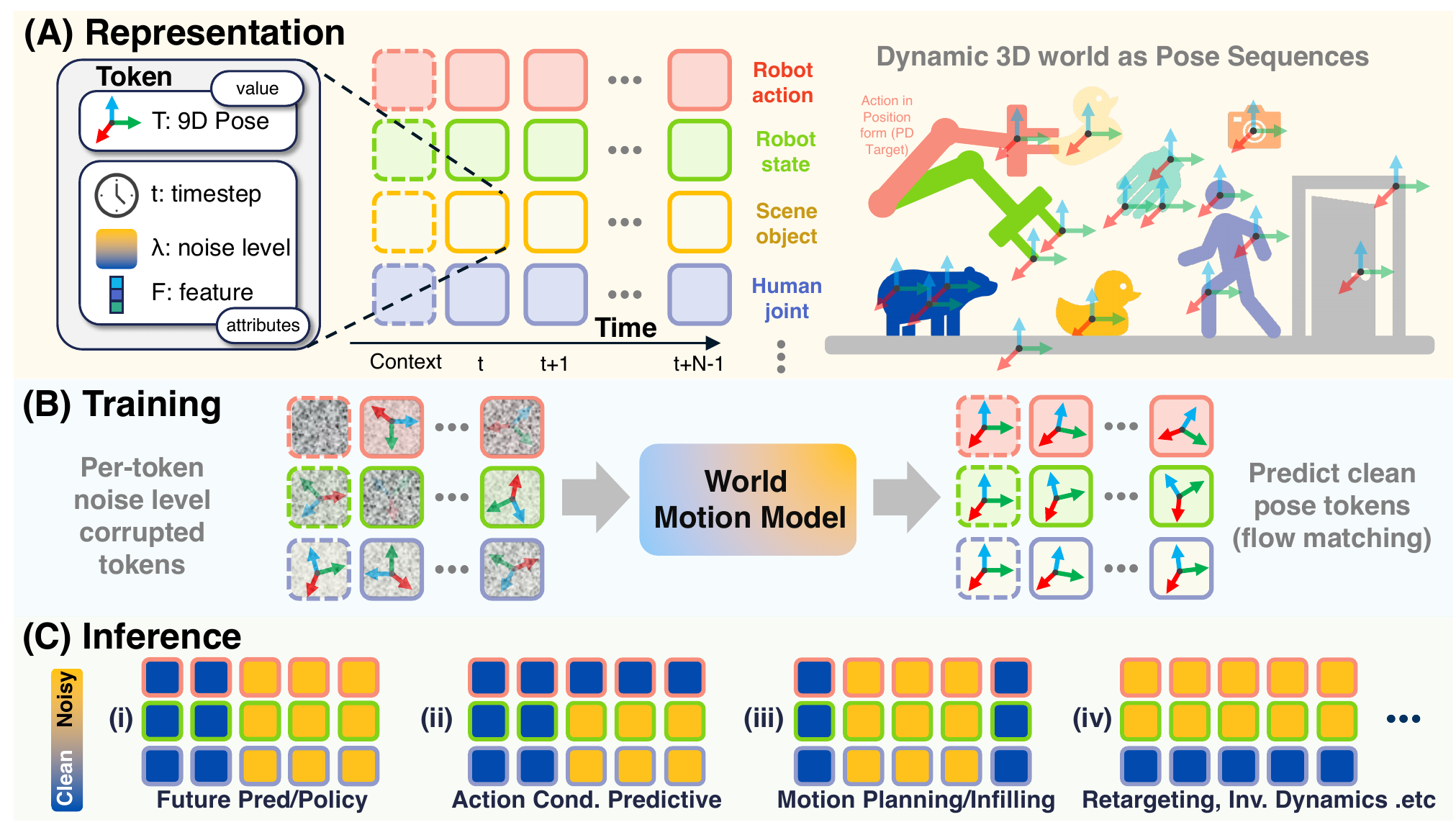}
    \par\vspace{-6pt}
    \caption{\textbf{World Motion Models}: (A) Dynamic scenes can be approximated by pose sequences. (B) During training, tokens corrupted at per-token noise levels are denoised to clean tokens. (C) During inference, various conditioning masks can be applied to one trained model.} %
    \label{fig:method:modes}

    \par\vspace{-10pt}
\end{figure}
\subsection{Representation: A Language for Motion Modeling}
\label{sec:method:representation}

\ourslong aim to answer \textbf{what was, is, and will be where} across time.
The thread connecting an entity at $t_1$ with the same entity at $t_2$ is its motion, and motion in real-world scenes is largely low-rank: a complex scene is driven by a relatively small set of \emph{eigen} rigid trajectories, while dense spatial deformation arises as a local combination of those bases~\cite{kratimenos2024dynmf,wang2024som,lei2024mosca}.
We assume the dynamic 3D world can be approximated by a finite set of rigid $\mathrm{SE}(3)$ trajectories, with dense spatial deformation recovered by interpolation, skinning, or binary assignment. $\mathrm{SE}(3)$ pose trajectories serve as the language with which we model the dynamic world. We %
leave modeling general linear local motion (scaling, shearing) to future work.

\vspace{-4pt}
\paragraph{Coverage.} \looseness=-1 As Fig.~\ref{fig:method:modes}-(A) shows, this primitive approximates the motion of the principal subjects of 3D vision and robotics. \emph{Scene future prediction} over 3D point trajectories~\cite{lei2025momaps,lee2025tracegen,huang2026pointworld} is the rotation-free special case of $\mathrm{SE}(3)$ trajectories. \emph{Human and hand-object motion} fits naturally: parametric body and hand models~\cite{loper2015smpl,romero2017mano,zuffi2017smal,lei2024gart} are sets of bone-joint $\mathrm{SE}(3)$ frames, and articulated parts contribute additional rigid trajectories with per-part local shape. \emph{Robot kinematics} treats each link as one rigid trajectory plus its mesh, the substrate for motion planning, retargeting, and control. \emph{Robot state and action} are equally expressible whenever the controller (PD gains, etc.) is fixed, since the action then reduces to a set of virtual target poses. Thus, the same language captures both the observation and action of a standard control problem. \emph{Camera motion} is, by definition, an $\mathrm{SE}(3)$ trajectory of the viewpoint.

\vspace{-4pt}
\paragraph{A pose at a time as a token.}
The elementary unit of our representation is a continuous token of ``where'', given by a pose $T^{(l)}\in SE(3)$ and paired with a feature $F^{(l)}\in \mathbb{R}^D$ that supplies the ``what'' at time $l$. From now on we use $\{i,j,k\}$ to index entities and $\{l,m,n\}$ to index successive time steps.
The feature is constant across time (shared along the trajectory) and is composed as $F_i=[\,f_i^{\text{sem}};\,f_i^{\text{geo}}\,]$: the semantic part $f_i^{\text{sem}}$ is either a DINO pixel feature~\cite{oquab2023dinov2} for video-prompted entities or a learnable embedding for coarse semantic categories (e.g.\ a body joint vs.\ an object). The geometric part $f_i^{\text{geo}}$ is a small descriptor extracted by a PointNet~\cite{qi2017pointnet} over the local point cloud when available (e.g.\ the local cloud of the object in human-object interaction or the shape of a robot body link) and is omitted otherwise (e.g.\ for a point prompt from a video). More details and ablations are in Appendix~\ref{app:more-abl-feature}.
The pose token $T_i^{(l)}$ itself is parameterized as a $9$-vector built from three points in $\mathbb{R}^3$,
\begin{equation}
    T_i^{(l)} \;\equiv\; \big(\,\mathbf{x},\;\;\mathbf{x}+\delta\,\mathbf{r}_1,\;\;\mathbf{x}+\delta\,\mathbf{r}_2\,\big)\in\mathbb{R}^9,
\end{equation}
where $\mathbf{x}\in\mathbb{R}^3$ is the position, $(\mathbf{r}_1,\mathbf{r}_2)$ are the first two columns of the rotation matrix under the 6D parameterization of~\citet{zhou2019continuity}, and $\delta$ is a small fixed scalar (e.g.\ $0.2$) empirically chosen to normalize the primal diffusion signal to a standard normal distribution (Appendix~\ref{app:more-abl-rotation-delta}). 
This $9$-D form folds the unit-vector pair $(\mathbf{r}_1,\mathbf{r}_2)$ into the same $\mathbb{R}^3$ space as the translation, making normalization and training easier.
Ground-truth labels for these poses at different time steps are widely available: they are read directly from parametric human reconstructions~\cite{goel2023humans4d,pavlakos2024hamer} or from robot action recordings. For real videos, they can be computed using a closed-form approximation by defining a proper neighborhood around point traces produced by recent 4D reconstruction tools~\cite{lei2025momaps,lee2025tracegen,huang2026pointworld}.

\vspace{-4pt}
\paragraph{Dynamic scenes as trajectory sequence sets.} \looseness=-1
An entity $i$ moving through time produces a sequence of such pose tokens decorated with the same feature,
\begin{equation}
    \tau_i \;=\; \big(\,[\,T_i^{(l)},\, T_i^{(m)},\, T_i^{(n)},\,\ldots\,];\; F_i\,\big),
\end{equation}
and a dynamic 3D scene is the set of $M$ such trajectories,
\begin{equation}
    \mathcal{X} \;=\; \{\tau_i,\,\tau_j,\,\tau_k,\,\ldots\}
    \;=\; \big\{\,([T_i^{(l)},T_i^{(m)},T_i^{(n)},\ldots]; F_i)\big\}_{i=1}^{M},
\end{equation}
which we abbreviate as $\mathcal{X}=(\mathcal{T},\mathcal{F})$ with the union of pose-sequences $\mathcal{T}=\{[T_i^{(l)},T_i^{(m)},T_i^{(n)},\ldots]\}_{i=1}^{M}$ and the corresponding feature set $\mathcal{F}=\{F_i,F_j,F_k,\ldots\}$.
Any dynamic scene over a causal temporal window can be expressed as such a pose-sequence set.
However, many problems also benefit from non-causal context, e.g.\ a goal state far beyond the rollout window for an MPC controller to optimize toward, or a handful of motion-pair exemplars that specify a cross-embodiment retargeting relationship between sequences. We therefore extend each trajectory with a prefix of virtual time steps disjoint from the causal window,
\begin{equation}
    \tau_i \;=\; \big([\,\tilde T_i^{(a)},\,\tilde T_i^{(b)},\,\ldots\;\big|\;\, T_i^{(l)},\,T_i^{(m)},\,T_i^{(n)},\,\ldots\,];\; F_i\big),
\end{equation}
where $a,b,\ldots$ are virtual indices distinguished by a separate positional embedding (Sec.~\ref{sec:method:arch}). Context tokens participate in training and inference exactly as causal ones, so a goal, an exemplar, or a label condition is simply another set of tokens. Without loss of generality, $\tau$ refers to this augmented form throughout, and Sec.~\ref{sec:exp} instantiates several choices of what to place in the virtual context.

\vspace{-4pt}
\paragraph{Modeling target.} A vanilla generative prior over this language captures $P\big(\,\mathcal{T} \,\big|\, \mathcal{F},\, \mathcal{A}\,\big)$, where $\mathcal{A}$ holds auxiliary conditions such as language or image tokens and $\mathcal{F}$ holds the per-token ``what'' decoration. Real applications, however, condition on arbitrary partial state rather than on $\mathcal{F}$ and $\mathcal{A}$ alone, so we instead study the family of any-marginal conditionals
\begin{equation}
    P\big(\,\mathcal{T}_{\mathrm{unk}} \,\big|\, \mathcal{T}_{\mathrm{obs}},\, \mathcal{F},\, \mathcal{A}\,\big),
    \qquad \forall\, \mathcal{T}_{\mathrm{obs}} \subset \mathcal{T},\;\; \mathcal{T}_{\mathrm{unk}} = \mathcal{T}\setminus \mathcal{T}_{\mathrm{obs}},
\end{equation}
i.e., every distribution induced by partitioning the trajectory set into an observed subset and its complement. We treat this any-to-any property as the operational definition of a world model, and a single \ourssing supports it.

\vspace{-4pt}
\subsection{Training: Flow Matching with Per-Token Noise Level}
\label{sec:method:training}

We learn the any-marginal prior of Sec.~\ref{sec:method:representation} with a single flow-matching model under \emph{asynchronous} noise levels (Fig.~\ref{fig:method:modes}-(B)): each token (one entity at one time step) is assigned its own scalar level, sampled i.i.d. This is inspired by Diffusion Forcing~\cite{chen2024diffusionforcing} and diffusion large language models (dLLMs)~\cite{nie2025large,ye2024beyond}, which can model sequences in arbitrary order and over arbitrary subsets.

\paragraph{Forward process.} Given a clean trajectory set $\mathcal{T}$ with $M$ entities and $N$ time steps (with pose tensor shape $M\times N\times 9$), we sample a per-token noise-level field $\Lambda\in[0,1]^{M\times N}$ with each entry $\lambda_{i,l}\sim\mathcal{U}[0,1]$ i.i.d.,
together with a Gaussian noise field $\epsilon\in\mathbb{R}^{M\times N\times 9}$, $\epsilon_{i,l}\sim\mathcal{N}(0,I_9)$. 
Each token moves along the path toward its sampled noise, with $\lambda_{i,l}$ broadcast across all $9$ channels, giving the noised pose
\begin{equation}
\label{eq:method:interp}
    T_i^{(l)}\Big|_{\lambda_{i,l}} \;=\; (1-\lambda_{i,l})\,T_i^{(l)} \;+\; \lambda_{i,l}\,\epsilon_{i,l}.
\end{equation}

\begin{figure}[!t]
    \centering
    \includegraphics[width=\linewidth]{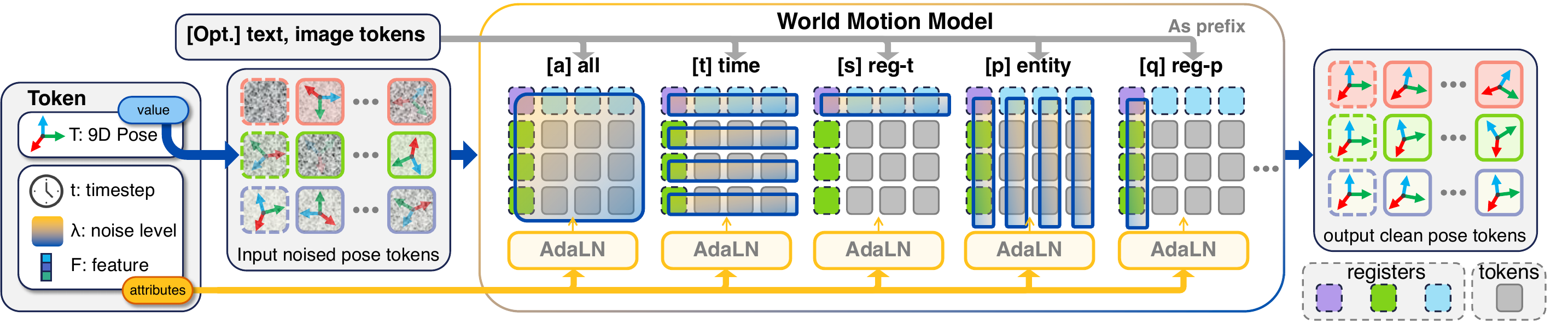}
    \par\vspace{-6pt}
    \caption{\textbf{Network}: The \ourssing architecture contains five factored attention types over a token grid padded with three register blocks, enabling efficient modeling (Sec.~\ref{sec:method:arch}).}
    \label{fig:method:arch}
    
    \par\vspace{-10pt}
\end{figure}

\paragraph{Velocity target and loss.} Our model $\mathcal{M}_{\theta}$ is a velocity field in the flow-matching sense~\cite{lipman2022flowmatching,liu2022rectified} that predicts the velocity from the clean pose toward its sampled noise. The interpolation target is $v^{\star}_{i,l} \;=\; \epsilon_{i,l} - T_i^{(l)}$,
and, conditioned on the noised trajectory from Eq.~\ref{eq:method:interp}, the feature set $\mathcal{F}$, the auxiliary condition $\mathcal{A}$, and the noise-level field $\Lambda$, the network is trained with a per-token error,
\begin{equation}
\label{eq:method:loss}
    \mathcal{L}(\theta) \;=\; \mathbb{E}_{\,\mathcal{T},\,\Lambda,\,\epsilon}\;\sum_{i,l}\;
    \Big\|\,\mathcal{M}_{\theta}\!\big(\mathcal{T}_{\Lambda},\,\mathcal{F},\,\mathcal{A},\,\Lambda\big)_{i,l} \;-\; v^{\star}_{i,l}\,\Big\|_{2}^{2}.
\end{equation}

\vspace{-4pt}
\paragraph{Why this single objective is enough.} Eq.~\eqref{eq:method:loss} bakes the any-marginal prior of Sec.~\ref{sec:method:representation} into one network~\cite{chen2024diffusionforcing,ye2024beyond}: each random $\Lambda$ splits tokens into nearly-clean (observation-like) and nearly-noised (target-like), covering every conditional reverse process, with the deterministic observed/target split as the limiting case. Fixing $\Lambda$ at inference (Sec.~\ref{sec:method:inference}) samples the chosen marginal. The noise level scheduling can also be biased toward deployment patterns, e.g.\ more often scheduling a small $\lambda$ on history and a large one on the future to sharpen forecasting while preserving coverage during training.

\vspace{-4pt}
\subsection{Inference: Various Marginal Distributions from the Same \ours}
\label{sec:method:inference}

A trained \ourssing is an entire family of conditional samplers under one set of weights. At inference, the user picks a per-token mask of observed and target tokens, sets $\lambda_{i,l}=0$ on the former and $\lambda_{i,l}=1$ on the latter, and integrates $\mathcal{M}_{\theta}$ from $\lambda=1$ down to $0$: observed tokens stay as input, while target tokens are transported by the velocity field, yielding a sample from $P(\mathcal{T}_{\mathrm{unk}}\mid\mathcal{T}_{\mathrm{obs}},\mathcal{F},\mathcal{A})$. Different masks cover various applications (Fig.~\ref{fig:method:modes}-C), for example:

\begin{itemize}[nosep,leftmargin=1.2em,label={}]
    \item \textbf{(i) Future prediction and policy.} $\lambda_{i,l}=0$ for $l\le l_0$, $\lambda_{i,l}=1$ for $l>l_0$. State-only entities give future prediction; with action entities in $\mathcal{T}$, the same mask is a robot policy.
    \item \textbf{(ii) Action-conditioned predictive model (forward dynamics).} Action entities known, future states unknown: a learned forward simulator for model-predictive control.
    \item \textbf{(iii) Motion planning and in-filling.} Sparse keyframes (possibly partial) are observed; the rest are targets: robot motion planning and motion interpolation.
    \item \textbf{(iv) Retargeting, partner prediction, and inverse dynamics.} Partition by entity: $\lambda_{i,l}=0$ for $i\in\mathcal{O}$, $\lambda_{i,l}=1$ otherwise. Taking $\mathcal{O}$ to be one body gives cross-embodiment retargeting; taking it to be a hand gives hand-object HOI partner prediction (or vice versa); taking it to be the state entities gives inverse dynamics.
\end{itemize}

Note that any known--unknown pattern is supported, beyond the examples mentioned above. Such a design has additional benefits: \textbf{robustness to observation noise.} Real observations are imperfect, and \ours accommodate this by setting $\lambda_{i,l}$ on observed tokens to a small positive value (e.g.\ $0.05$) rather than exactly $0$, which turns each observation into a strong hint and lets the integrator slightly rectify the token along the velocity field as the noise level decreases. \textbf{Generalization to many sequences:} RoPE in a transformer already helps the model generalize beyond the number of sequences seen during training. A simple inference-time algorithm (App.~\ref{app:dense-sampler}) extends this generalization further when combined with the per-token noise formulation, enabling dense trajectory generation well beyond the training count.

\subsection{Implementation: Model Architecture}
\label{sec:method:arch}
As shown in Fig.~\ref{fig:method:arch}, our model supports arbitrary numbers of entities and time steps while addressing three issues: (i) \emph{per-token noise levels}: $\lambda_{i,l}$ has to be specified \emph{per token}, not as a single global scalar as in a conventional flow-matching backbone~\cite{peebles2023dit,lipman2022flowmatching}; (ii) \emph{scale}: a modest $30$-frame, $400$-trajectory scene already contains $12$K tokens, on which dense self-attention is expensive; (iii) \emph{missing observations}: real data come with invalid intervals (occlusion, truncation, tracker drop-out) that cannot simply be masked during training. We address each as follows:
\textbf{(1) Per-token conditioning via AdaLN.} For each token $(i,l)$ we concatenate (a) a sinusoidal embedding of $\lambda_{i,l}$, (b) the feature $F$, and (c) a pooled embedding of the auxiliary tokens $\mathcal{A}$ (e.g.\ a language summary). A per-block wide multi-layer FFN maps this vector to per-token AdaLN scale-and-shift parameters that modulate every transformer block~\cite{peebles2023dit}.
\textbf{(2) Registers and factored attention.} We pad the $M\times N$ token grid with three blocks of learnable registers~\cite{darcet2024registers} and apply five attention types on the padded grid: \texttt{a} self-attends across all tokens; \texttt{t} and \texttt{p} are time- and entity-axis self-attention; \texttt{s} and \texttt{q} are register-only counterparts on the same two axes, widened aggressively in channels at negligible runtime cost to add expressivity.
\textbf{(3) Training under missing observations.} Invalid time steps must still participate in attention, since at inference the missing mask is unknown. We pin invalid tokens to $\lambda_{i,l}=1$ (pure noise) and sum the loss in Eq.~\eqref{eq:method:loss} only over tokens with ground-truth labels.
Virtual-context tokens (Sec.~\ref{sec:method:representation}) are flagged by a $1$-bit prefix mask and given learnable position embeddings. Text or image tokens attach as an attention prefix.
\vspace{-4pt}
\section{Experiments}
\label{sec:exp}

We evaluate \ours on open-source benchmarks spanning policy learning, future prediction, HOI, and retargeting, plus a set of ablations. 
We train a separate \ourssing checkpoint on each benchmark's training split; joint training of one \ourssing across all benchmarks is left to future work.
The entities modeled by a \ourssing are defined according to the task: examples include a link or joint of a robot, human body, or object 
(Secs.~\ref{sec:exp:lt}, \ref{sec:exp:hh}, and \ref{sec:exp:om}), or any prompted query point in the general background (Sec.~\ref{sec:exp:tg}); see below for details. 

\begin{figure}[!t]
    \centering
    \includegraphics[width=\linewidth]{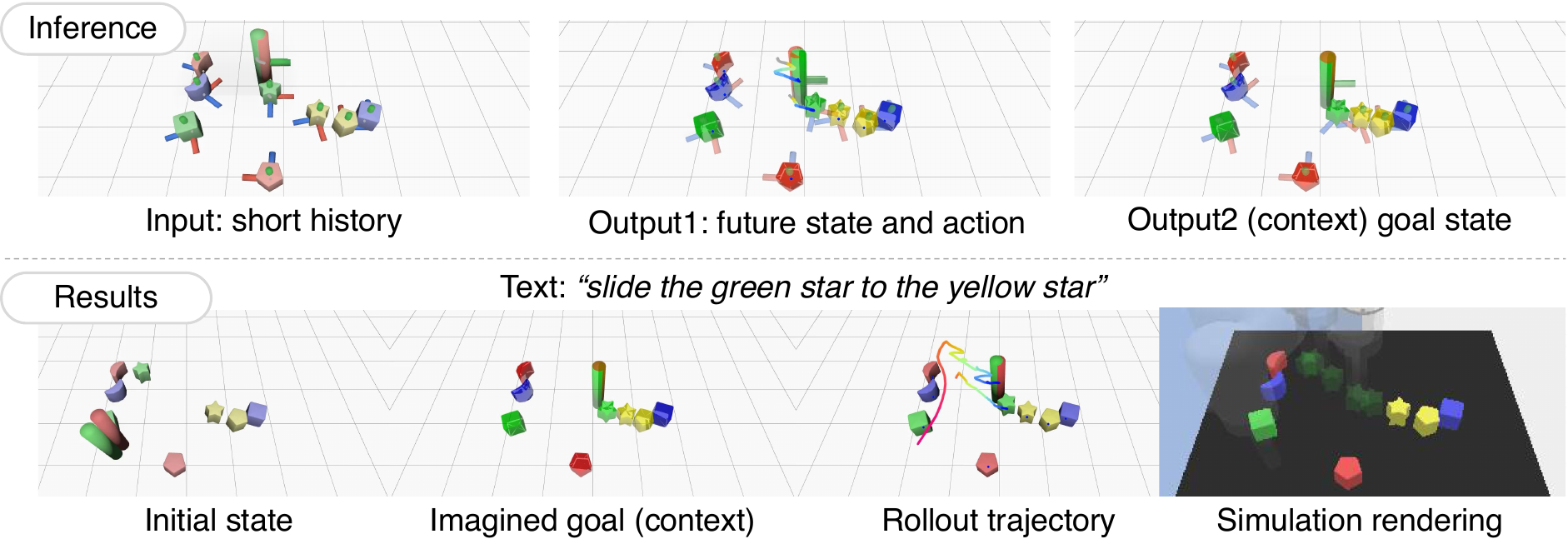}
    \par\vspace{-3pt}
    \caption{\textbf{Language Table.} Top: inference pattern; history is the input, while future and goal are predicted by \ours. Bottom: policy rollout results.}
    \label{fig:exp:lt}
    \par\vspace{2pt}
\begin{minipage}{\linewidth}
\captionsetup{type=table}
\centering
\begin{minipage}[t]{0.66\textwidth}
\centering
\caption{\textbf{Language Table.} Per-task and macro-average success rate (\%).}
\label{tab:exp:lt}
\small
\setlength{\tabcolsep}{2.5pt}
\begin{tabular}{lrrrrrr}
\toprule
Method & B2B$\uparrow$ & B2AL$\uparrow$ & B2BRL$\uparrow$ & B2RL$\uparrow$ & Sep$\uparrow$ & Avg$\uparrow$ \\
\midrule
Scratch~\cite{ye2024lapa}                            &  4.0 &  6.0 & 10.0 &  6.0 & 52.0 & 15.6 \\
UniPi~\cite{du2023unipi}                                & 14.0 &  4.0 & 12.0 & 10.0 & 72.0 & 22.4 \\
VPT~\cite{baker2022vpt}                                 & 36.0 & 38.0 & 48.0 & 26.0 & 70.0 & 43.6 \\
LAPA~\cite{ye2024lapa}                               & 58.0 & 56.0 & 52.0 & 48.0 & 96.0 & 62.0 \\
ActionVLA~\cite{ye2024lapa}                          & 76.0 & 72.0 & 76.0 & 70.0 & 90.0 & 76.8 \\
LCB~\cite{shentu2024lcb}                                & \textbf{90.0} & 71.0 & 66.0 & 73.0 & 99.0 & 80.0 \\
LangTable orig.~\cite{lynch2022langtable}               & 88.0 & 65.0 & 70.0 & 68.0 & 94.0 & 77.0 \\
\textbf{\ourssing}                                          & \textbf{90.0} & \textbf{76.0} & \textbf{80.0} & \textbf{90.0} & \textbf{100.0} & \textbf{87.2} \\
\bottomrule
\end{tabular}
\end{minipage}\hfill
\begin{minipage}[t]{0.32\textwidth}
\centering
\caption{\textbf{Plain Policy versus MPC.} On Language Table absolute tasks.}
\label{tab:exp:lt:mpc}
\footnotesize
\setlength{\tabcolsep}{1.5pt}
\begin{tabular}{l rr rr}
\toprule
 & \multicolumn{2}{c}{B2B} & \multicolumn{2}{c}{B2AL} \\
\cmidrule(lr){2-3}\cmidrule(lr){4-5}
 & Plain & MPC & Plain & MPC \\
\midrule
Succ.$\uparrow$           & 90.0 & \textbf{100.0} & 76.0 & \textbf{80.0} \\
Mean steps$\downarrow$    & 79.5 & \textbf{79.1}  & 92.7 & \textbf{87.2} \\
Med.\ steps$\downarrow$   & 73.0 & \textbf{68.5}  & 87.0 & \textbf{85.5} \\
Min steps$\downarrow$     & 26   & \textbf{16}    & 14   & \textbf{2}    \\
Max steps$\downarrow$     & 198  & \textbf{175}   & 187  & \textbf{171}  \\
\bottomrule
\end{tabular}
\end{minipage}
\end{minipage}

    \par\vspace{-10pt}
\end{figure}

\vspace{-4pt}
\subsection{Language Table: Policy Learning and Forward Dynamics} 
\label{sec:exp:lt}
Language Table~\cite{lynch2022langtable} is a tabletop pushing benchmark with five task families.
We follow the evaluation protocol of LAPA~\cite{ye2024lapa} and LCB~\cite{shentu2024lcb}:
$50$ rollouts per family on the eight-block setup with a $200$-step horizon, reporting per-family and macro-average success rates.
\textbf{Setup.} As in Fig.~\ref{fig:exp:lt}, the entity set comprises the 6D pose of every block together with the robot end-effector state (green) and action target (red); the context tokens are configured to capture the final frame of a successful trajectory.
The feature descriptor of each token is defined as follows: $F_{\mathrm{sem}}$ consists of three learnable embeddings for object, link-state, and link-action, while $F_{\mathrm{geo}}$ is obtained by applying a PointNet to the local point cloud of the object or link.
We train one model on the public oracle trajectories across all five tasks, and inference runs in the policy mode of Sec.~\ref{sec:method:inference}: given a five-step history and language prompt, a ten-step action chunk is read out from the action entity in the future prediction, ensembled across queries as in ACT~\cite{zhao2023act}. The context (goal state) is also generated from the language and observation.
\textbf{Results.} As in Tab.~\ref{tab:exp:lt} and Fig.~\ref{fig:exp:lt}, a \ourssing obtains the strongest macro-average, demonstrating that \ours are effective and accurate as policies.

\vspace{-4pt}
\paragraph{Action optimization with MPC.} \ours can also be used as a differentiable world model rather than a plain policy. For tasks with absolute-position instructions (B2B, B2AL) in LangTable, given the generated goal state (via context tokens), we can apply model predictive control (MPC) to optimize the predicted action so as to maximize a heuristically defined value function: the similarity between the final frame of the action chunk and the generated goal. Since our model is fully differentiable, gradient-based MPC can be hybridized with sampling-based methods (full procedure in App.~\ref{app:exp:lt:mpc}); Tab.~\ref{tab:exp:lt:mpc} compares the plain policy with the MPC-optimized policy on B2B and B2AL, indicating both a higher success rate and faster rollout to success.

\begin{figure}[!t]
    \centering
    \includegraphics[width=\linewidth]{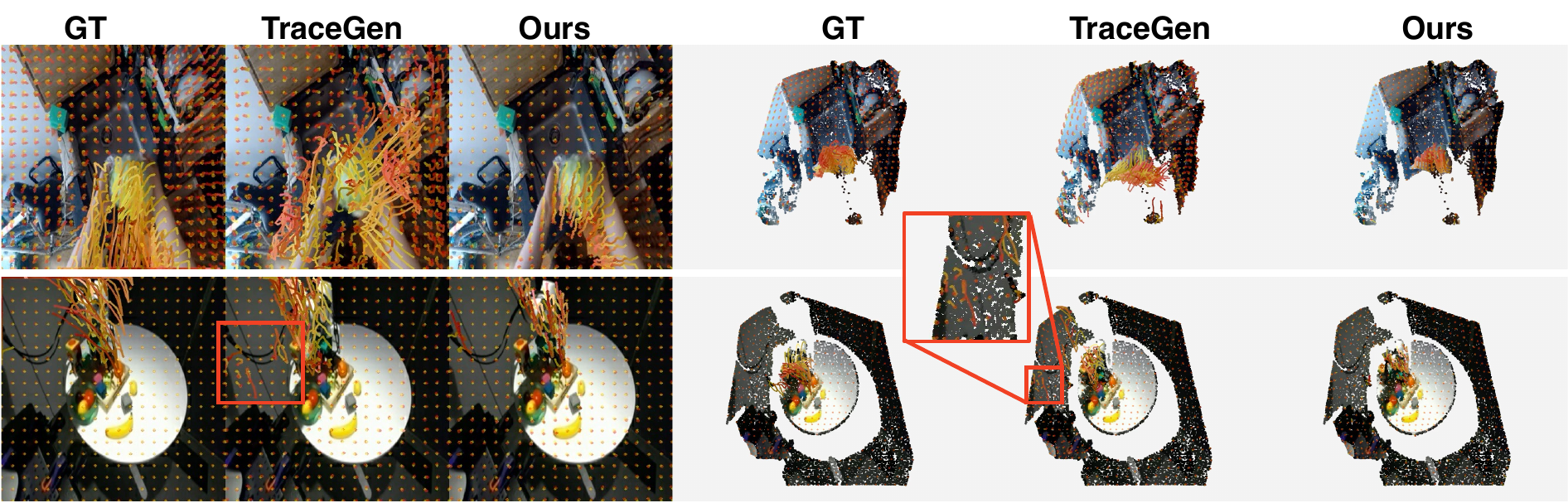}
    \par\vspace{-6pt}
    \caption{\textbf{TraceGen.} Qualitative results on the TraceGen benchmark~\cite{lee2025tracegen}.}
    \label{fig:exp:tg}
    \par\vspace{4pt}
\begin{minipage}{\linewidth}
\captionsetup{type=table}
\centering
\caption{\textbf{TraceGen.} 3D-trajectory metrics on the Droid and EpicKitchen test splits.}
\label{tab:exp:tg}
\small
\setlength{\tabcolsep}{3.5pt}
\begin{tabular}{l|rrr|rrr}
\toprule
Method & Dr MSE$\downarrow$ & Dr MAE$\downarrow$ & Dr Endp.$\downarrow$ & Ep MSE$\downarrow$ & Ep MAE$\downarrow$ & Ep Endp.$\downarrow$ \\
\midrule
TraceGen~\cite{lee2025tracegen}                        & 0.206          & 1.289          & 0.285          & 0.445          & 2.721          & 0.791          \\
\midrule
\ourssing, per-dataset ($P{=}128$)                             & \textbf{0.151} & \textbf{1.129} & \textbf{0.200} & \textbf{0.322} & \textbf{2.130} & \textbf{0.556} \\
\ourssing, per-dataset ($P{=}400$)                             & 0.178          & 1.169          & 0.236          & 0.329          & 2.167          & 0.580          \\
\ourssing-Joint ($P{=}128$)                                & 0.167          & 1.172          & 0.225          & 0.346          & 2.256          & 0.615          \\
\ourssing-Joint ($P{=}400$)                                & 0.187          & 1.179          & 0.259          & 0.351          & 2.287          & 0.621          \\
\bottomrule
\end{tabular}

\end{minipage}

    \par\vspace{-10pt}
\end{figure}

\vspace{-4pt}
\subsection{TraceGen: Dense 3D Scene Future Prediction} 
\label{sec:exp:tg}
TraceGen~\cite{lee2025tracegen} is a 3D trajectory generation benchmark in which the model takes a single RGB-D start frame and a language instruction and predicts the motion of $400$ keypoints over $32$ steps. We follow its protocol on the two subsets that provide 3D trajectories, Droid and EpicKitchen, reporting the trajectory-generation errors MSE, MAE, and endpoint MSE (each $\times 100$).
\textbf{Setup.} Each of the $400$ keypoints is one entity carrying a 3D position and an orientation (computed by local Procrustes in the dataset), starting at the first frame ($t{=}0$). We train four \ourssing checkpoints over the experimental space (Droid only, EpicKitchen only, and joint on both) at two keypoint window sizes ($P{=}128$ and $P{=}400$, where $P$ is the number of trajectories seen during training). No special context token is used in this experiment.
\textbf{Results.} As shown in Tab.~\ref{tab:exp:tg} and Fig.~\ref{fig:exp:tg}, \ourssing achieves the best performance in every column in both environments.
Note that the $P=128$ variant is trained with $128$ randomly selected trajectories, meaning that it can accept random queries during inference.
Although only $128$ trajectories are observed in parallel during training, the model can directly infer all $400$ trajectories and even achieves better performance (P=128 wins all metrics); we hypothesize that this stems from the random sampling of $128$ out of $400$ trajectories during training, which prevents overfitting. The experiments also show that joint training across both domains is feasible, suggesting that future work can pursue a single unified model across domains.

\vspace{-4pt}
\subsection{OMOMO: Human-Object Interaction}
\begin{figure}[!t]
    \centering
    \includegraphics[width=\linewidth]{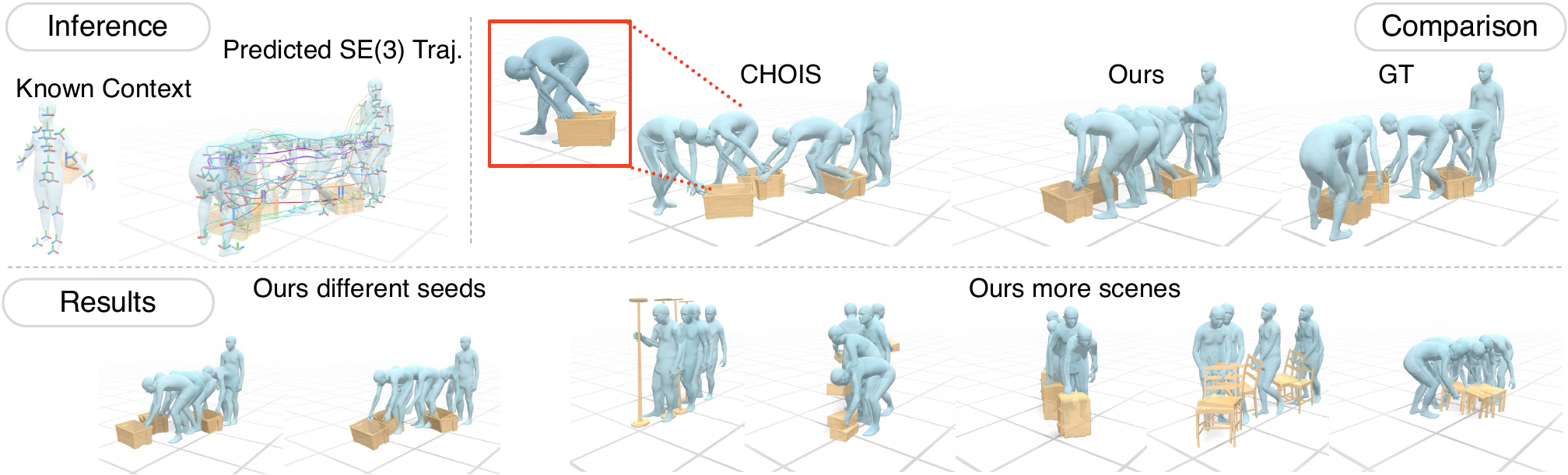}
    \par\vspace{-4pt}
    \caption{\textbf{OMOMO.} Qualitative results on OMOMO~\cite{li2023omomo}. Top left: inference pattern. Top right: comparison with CHOIS~\cite{li2024chois}. Bottom: additional results from \ours.}
    \label{fig:exp:om}
    \par\vspace{2pt}
\begin{minipage}{\linewidth}
\captionsetup{type=table}
\centering
\caption{\textbf{OMOMO.} Results under the CHOIS~\cite{li2024chois} protocol.}
\label{tab:exp:om}
\footnotesize
\setlength{\tabcolsep}{2.2pt}
\resizebox{\linewidth}{!}{%
\begin{tabular}{lrrrrrrrrrrrrr}
\toprule
Method & T$_{s}\!\downarrow$ & T$_{e}\!\downarrow$ & T$_{xy}\!\downarrow$ & H$_{\text{feet}}\!\downarrow$ & FS$\downarrow$ & C$_{\text{prec}}\!\uparrow$ & C$_{\text{rec}}\!\uparrow$ & C$_{\text{F1}}\!\uparrow$ & C$_{\%}\!\uparrow$ & MPJPE$\downarrow$ & T$_{\text{root}}\!\downarrow$ & T$_{\text{obj}}\!\downarrow$ & O$_{\text{obj}}\!\downarrow$ \\
\midrule
InterDiff~\cite{xu2023interdiff}                    & --   & 158.84 & 72.72 & 0.90 & 0.42 & 0.63 & 0.28 & 0.33 & 0.27 & 25.91 & 63.44 & 88.35 & 1.65 \\
MDM~\cite{tevet2022mdm}                             & 5.18 & 33.07 & 19.42 & 6.72 & 0.48 & 0.72 & 0.47 & 0.53 & 0.43 & 17.86 & 34.16 & 24.46 & 1.85 \\
Pred-OMOMO~\cite{li2023omomo}                        & 2.39 & 8.03 & 4.15 & 7.08 & 0.40 & 0.73 & 0.66 & 0.66 & \textbf{0.62} & 18.66 & 28.39 & 16.36 & 1.05 \\
CHOIS w/o L$_{obj}$~\cite{li2024chois}               & 5.76 & 14.16 & 8.44 & 6.55 & 0.40 & 0.75 & 0.50 & 0.55 & 0.43 & 14.34 & 21.97 & 15.53 & \textbf{0.98} \\
CHOIS w/o F$_{all}$~\cite{li2024chois}               & 1.75 & 6.61 & \textbf{2.69} & 6.64 & 0.38 & 0.78 & 0.49 & 0.55 & 0.41 & 15.23 & 24.13 & \textbf{11.51} & 0.99 \\
CHOIS~\cite{li2024chois}                             & 1.71 & \textbf{6.31} & 2.87 & 4.20 & \textbf{0.35} & \textbf{0.80} & 0.64 & 0.67 & 0.54 & 15.30 & 24.43 & 12.53 & 0.99 \\
\textbf{\ourssing}                                       & \textbf{0.15} & 6.48 & 4.40 & \textbf{0.84} & 0.56 & 0.79 & \textbf{0.67} & \textbf{0.76} & 0.59 & \textbf{12.07} & \textbf{19.87} & 16.53 & 1.03 \\
\bottomrule
\end{tabular}%
}

\end{minipage}

    \par\vspace{-10pt}
\end{figure}
\label{sec:exp:om}
OMOMO~\cite{li2023omomo} is a benchmark for object-conditioned full-body human-object interaction synthesis: given a partial object keyframe and a language prompt, the model generates the body and object motion. We follow the protocol of CHOIS~\cite{li2024chois}.
\textbf{Setup.} The entity set comprises the SMPL body joints and the interaction object's rigid poses. The Context tokens are always set to a SMPL human in an A-pose (which conveys entity identities) and the object at the origin. The time window is set to the dataset's full length of $120$ frames.

Note that in this human body example and the following humanoid and hand experiments, \ourssing directly models pose trajectories even when a structure such as the SMPL model is known, enabling \ourssing to model general and diverse morphologies within a single architecture and model. The following visualizations are produced by directly animating the meshes with linear blend skinning (LBS) or rigid transformations using the poses predicted by \ourssing, showing the accuracy of our model.
To conduct a fair comparison (e.g., baselines use SMPL parameters to compute metrics), we convert the output of \ourssing to the corresponding parametric space; further discussion of this projection is provided in Appendix~\ref{app:more-abl-projection}.

\textbf{Results.} Tab.~\ref{tab:exp:om} and Fig.~\ref{fig:exp:om} show comparable or even better performance on the HOI synthesis task. Fig.~\ref{fig:exp:om} further shows that when CHOIS produces implausible predictions in which the human intersects the object, the \ourssing maintains plausibility and can generate diverse motions.
Additionally, to compare model size and efficiency, we measure the parameter counts of TraceGen and \ourssing (674.6M vs 438.8M) and their inference times for 400 tracks over 32 steps (2.66 s vs 4.18 s).

\vspace{-4pt}
\subsection{Human and Humanoid: Kinematic Retargeting}
\label{sec:exp:hh}
\begin{figure}[!t]
    \centering
    \includegraphics[width=\linewidth]{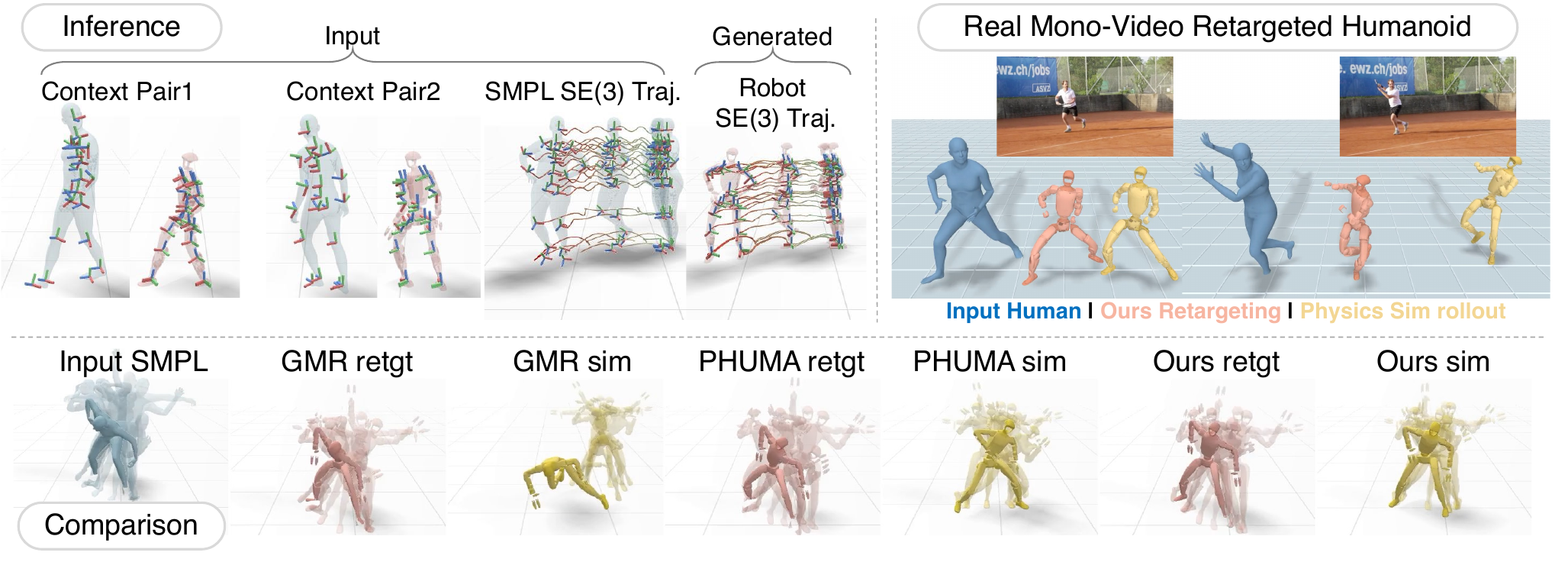}
    \par\vspace{-6pt}
    \caption{\textbf{Human-to-Humanoid Retargeting.} Top left: inference pattern. Top right: retargeting from real-world video. Bottom: comparison against baselines.}
    \label{fig:exp:hh}
    \par\vspace{4pt}
\begin{minipage}{\linewidth}
\captionsetup{type=table}
\centering
\caption{\textbf{Humanoid Retargeting.} Results under three conditioning regimes (clean, sparse-keyframe in-fill, per-frame noised).}
\label{tab:exp:hh}
\footnotesize
\setlength{\tabcolsep}{4pt}
\resizebox{\linewidth}{!}{%
\begin{tabular}{l r r r r r r r r r r}
\toprule
 & \multicolumn{4}{c}{Clean (30\,FPS)} & \multicolumn{3}{c}{In-fill (3\,FPS keyframes)} & \multicolumn{3}{c}{Noisy ($\sigma{=}0.1$)} \\
\cmidrule(lr){2-5} \cmidrule(lr){6-8} \cmidrule(lr){9-11}
Method & FPS$\uparrow$ & SR-seq$\uparrow$ & gerr$\downarrow$ & lerr$\downarrow$ & SR-seq$\uparrow$ & gerr$\downarrow$ & lerr$\downarrow$ & SR-seq$\uparrow$ & gerr$\downarrow$ & lerr$\downarrow$ \\
\midrule
GMR~\cite{araujo2025gmr}         & 35.03  & 0.7945 & 0.6911 & 0.0651 & 0.8095 & 0.6542 & 0.0592 & 0.7160 & 1.0196 & 0.1158 \\
PHUMA~\cite{lee2025phuma}        & 2.27   & 0.8255 & \textbf{0.5509} & 0.0530 & 0.6535 & 1.3744 & 0.0745 & 0.8060 & \textbf{0.7269} & 0.0703 \\
\textbf{\ourssing}                  & \textbf{86.97} & \textbf{0.8690} & 0.5938 & \textbf{0.0420} & \textbf{0.8525} & \textbf{0.6371} & \textbf{0.0518} & \textbf{0.8530} & 0.7785 & \textbf{0.0691} \\
\bottomrule
\end{tabular}%
}
\end{minipage}

    \par\vspace{-10pt}
\end{figure}

We construct a benchmark that takes a SMPL human motion sequence and produces a Unitree G1 humanoid~\cite{unitree2024g1} trajectory. We evaluate retargeting quality by feeding the trajectory to a low-level robot controller (TWIST~\cite{ze2025twist}), rolling it out in physics simulation, and measuring tracking error and success rate.
Three input settings probe \ours' flexibility: clean human input at $30$\,FPS; sparse keyframes at $3$\,FPS that the model must infill; and human input with noise at every frame ($\sigma{=}0.1$ on axis-angle and root translation).
\textbf{Setup.} The entity set comprises the SMPL human and the G1 humanoid links, paired across embodiments inside the Context-token exemplar frames (Sec.~\ref{sec:method:representation}; two samples in Fig.~\ref{fig:exp:hh}). A \ourssing is trained on a mix of AMASS~\cite{mahmood2019amass} and Embody3D~\cite{mclean2025embody3d}; a single checkpoint serves all three settings. The humanoid's kinematic reference (ground truth) is obtained by running and filtering PHUMA~\cite{lee2025phuma}, a slow offline optimization method.
\textbf{Results.} Tab.~\ref{tab:exp:hh} and Fig.~\ref{fig:exp:hh} report the comparison against GMR~\cite{araujo2025gmr} and PHUMA~\cite{lee2025phuma}. GMR is a fast, real-time method but provides limited quality; PHUMA is an offline optimization method that produces more physically plausible results (less overlap, etc.) but is slow. The resulting \ourssing is a retargeting prior that is fast (note the throughput reported in the first column) and learns a prior over physically plausible, safe motion across sequences, providing higher-quality and safer retargeting motion. Fig.~\ref{fig:exp:hh} also shows real-world monocular-video-based humanoid retargeting and control results.

\vspace{-4pt}
\subsection{More Applications} 
\label{sec:exp:more}
\begin{figure}[t]
    \centering
    \includegraphics[width=1.0\linewidth]{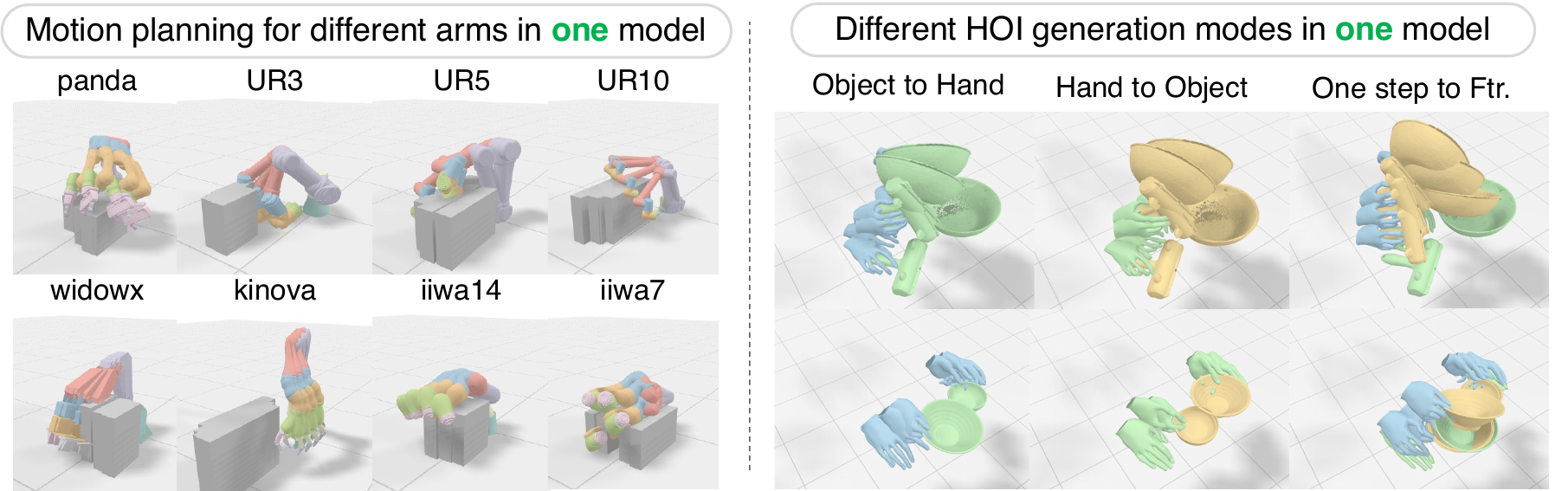}
    \par\vspace{-6pt}
    \caption{\textbf{More Applications.} Qualitative results of \ours on additional tasks.}
    \label{fig:exp:more}
    \par\vspace{-10pt}
\end{figure}

The left side of Fig.~\ref{fig:exp:more} shows an additional application of \ours: motion planning with known first and last timesteps (Fig.~\ref{fig:method:modes}-(C ii), labelled by PyRoKi~\cite{kim2025pyroki}). This works across more than 8 different manipulators within a single model, showing that our flexible modeling techniques can model diverse kinematic targets across morphologies.
The right side of Fig.~\ref{fig:exp:more} shows hand-object interaction generation on TACO~\cite{liu2024taco} under various known-unknown patterns within one model (the green mesh denotes the known condition), showing that \ourssing can be used in diverse ways during inference.

\vspace{-4pt}
\subsection{Ablations} 
\begin{wraptable}{R}{0.55\linewidth}
\centering
\vspace{-1.2em}
\caption{\textbf{Ablations.} Language Table B2B (left), TraceGen EpicKitchen (right).}
\label{tab:exp:abl}
\footnotesize
\setlength{\tabcolsep}{3pt}
\begin{tabular}{lr||lrrr}
\toprule
Variant          & SR$\uparrow$    & Variant              & MAE$\downarrow$ & MSE$\downarrow$ & Endp.$\downarrow$ \\
\midrule
\textbf{\ourssing}   & \textbf{0.860}  & \textbf{\ourssing}       & \textbf{2.167}  & \textbf{0.329}  & \textbf{0.580}    \\
\,\,no AdaLN     & 0.800           & \,\,no eff.          & 2.356           & 0.344           & 0.602             \\
\,\,no per-tk    & 0.000           & \,\,plain DiT        & 32.05           & 16.97           & 17.64             \\
\bottomrule
\end{tabular}
\vspace{-1em}
\end{wraptable}

\label{sec:exp:abl}
We ablate per-block AdaLN and the per-token-noise schedule (Sec.~\ref{sec:method:training}) on Language Table B2B, and architectural primitives (registers, factored axis-wise attention; Sec.~\ref{sec:method:arch}) on TraceGen EpicKitchen. Each ablation toggles one field; details in App.~\ref{app:exp:abl}.
\textbf{Language Table.} Tab.~\ref{tab:exp:abl} (left): removing AdaLN drops B2B success by $6$ pp but the model still learns a competent policy; removing per-token noise (using synced noise with guidance-based conditioning) yields no successful rollout, since policy mode requires very high accuracy.
\textbf{TraceGen.} Tab.~\ref{tab:exp:abl} (right): collapsing the middle three layers to plain DiT (no efficient register) degrades performance because the same hardware then supports a smaller batch size and slower training. Collapsing all seven to plain DiT (no axis-wise factorization) yields a $\sim 15\times$ MAE regression, as the model loses the explicit temporal--spatial inductive bias.

\vspace{-4pt}
\section{Related Work}
\label{sec:related_short}

\paragraph{Motion Generation and World Models.} 
Existing generative motion models are fragmented by actor class through morphology-specific parameterizations~\cite{tevet2022mdm,jiang2024motiongpt,li2026llamomotion}, while general unified approaches~\cite{lei2025momaps,huang2026pointworld,nvidia2025cosmos} operate in dense observation spaces such as pixel-aligned point tracks, 3D flow, or video entangling 3D dynamics with appearance, or solely in 2D pixel motion space~\cite{ranasinghe2025pixel,nguyen2026pixel,ranasinghe2026future}. A piecewise-rigid lineage uses $SE(3)$ transforms~\cite{byravan2017se3nets,hsu2024spot} but is restricted to single-agent or fixed-object settings. \ours{} instead encode every moving unit (robot link, body part, scene primitive) as a (sparse) set of $SE(3)$ tokens in a shared abstraction, leading to a potentially more unified and shared motion prior over all moving subjects in the physical world.

\vspace{-4pt}
\paragraph{Flexible Policies and Sequence Modeling.} 
Recent work couples world modeling with policy via video--action models~\cite{liu2026worldvla,zhu2025uwm,li2025uva} or latent-dynamics planners~\cite{hansen2024hwm,hafner2023dreamerv3}, but ties the world model to a fixed morphology or pixel tokens. 
Diffusion Forcing~\cite{chen2024diffusionforcing} and discrete masked diffusion~\cite{arriola2025blockdiffusion} introduce per-token noise schedules for flexible conditioning.
Specifically, Diffusion Forcing~\cite{chen2024diffusionforcing} is the modeling method closest to ours. Although theoretically general, it primarily focuses on temporal scheduling across sequence frames. Our method generalizes the idea of per-token noise to spatial and temporal axes, and our architectural design makes this practical for more than 10K tokens.
We provide a unified representation of heterogeneous SE(3) trajectories and add a decoupled context-token mechanism, so one network spans prediction, policy rollout, and non-causal planning. See App.~\ref{sec:related} for an extended discussion.

\vspace{-4pt}
\section{Conclusion}
\label{sec:conclusion}

We propose \ourslong, a unified generative prior over the dynamic 3D world via sparse $\mathrm{SE}(3)$ pose trajectories. A per-token-noise-level flow-matching model with context tokens supports any-marginal conditional inference from one network, attaining state-of-the-art results on policy learning, scene future prediction, human-object interaction, and humanoid retargeting. 

\vspace{-4pt}
\paragraph{Limitations.} \textbf{(i)} We train one \ourssing checkpoint per benchmark; joint training across domains and increasing the amount of supervision data are left to future work. 
\textbf{(ii)} Generalizing $SE(3)$ to more general linear groups is also a future direction. 
\textbf{(iii)} The current framework requires geometry (e.g., depth and estimated current and past states) as input; relaxing these requirements and enabling direct inference from RGB videos are also directions for future work. 
\textbf{(iv)} Another limitation of the current model is that it does not generalize to novel morphologies unseen during training; further investigation of in-context learning with our proposed context mechanism may suggest solutions in the future.
\par\vspace{-4pt}
\paragraph{Broader impact.} \ourssing may have benefits in robotics, mixed reality, animation, games, and simulation. We do not see major direct negative ethical or societal consequences of our work. 

\clearpage

\begin{ack}
This project was funded in part by Meta BAIR partners, the BAIR Humanoid Intelligence Center (BAIR HIC), and NSF CAREER (No. 2442491).
We used LLMs and agents to write and debug code, help review the literature, and polish the writing.
\end{ack}

\bibliographystyle{plainnat}
\bibliography{references}

@article{ranasinghe2025pixel,
  title={Pixel motion as universal representation for robot control},
  author={Ranasinghe, Kanchana and Li, Xiang and Nguyen, E-Ro and Mata, Cristina and Park, Jongwoo and Ryoo, Michael S},
  journal={arXiv preprint arXiv:2505.07817},
  year={2025}
}

@inproceedings{nguyen2026pixel,
  title={Pixel motion diffusion is what we need for robot control},
  author={Nguyen, E-Ro and Zhang, Yichi and Ranasinghe, Kanchana and Li, Xiang and Ryoo, Michael S},
  booktitle={Proceedings of the IEEE/CVF Conference on Computer Vision and Pattern Recognition},
  pages={23663--23672},
  year={2026}
}

@article{ranasinghe2026future,
  title={Future optical flow prediction improves robot control \& video generation},
  author={Ranasinghe, Kanchana and Zhou, Honglu and Fang, Yu and Yang, Luyu and Xue, Le and Xu, Ran and Xiong, Caiming and Savarese, Silvio and Ryoo, Michael S and Niebles, Juan Carlos},
  journal={arXiv preprint arXiv:2601.10781},
  year={2026}
}

@article{chen2026oasis,
  title={OASIS: Observation-Action Space Alignment via SE (3) Trajectory Prediction for Robotic Manipulation},
  author={Chen, Xinzhe and Ren, Sihua and Huang, Liqi and Sun, Haowen and Li, Mingyang and Chen, Xingyu and Liu, Zeyang and Lan, Xuguang},
  journal={arXiv preprint arXiv:2605.25829},
  year={2026}
}

@inproceedings{qi2017pointnet,
  title     = {Pointnet: Deep learning on point sets for 3d classification and segmentation},
  author    = {Qi, Charles R and Su, Hao and Mo, Kaichun and Guibas, Leonidas J},
  booktitle = {Proceedings of the IEEE conference on computer vision and pattern recognition},
  pages     = {652--660},
  year      = {2017}
}

@inproceedings{ye2024beyond,
  title     = {Beyond autoregression: Discrete diffusion for complex reasoning and planning},
  author    = {Ye, Jiacheng and Gao, Jiahui and Gong, Shansan and Zheng, Lin and Jiang, Xin and Li, Zhenguo and Kong, Lingpeng},
  booktitle = {International Conference on Learning Representations},
  volume    = {2025},
  pages     = {77875--77898},
  year      = {2025}
}

@article{nie2025large,
  title   = {Large language diffusion models},
  author  = {Nie, Shen and Zhu, Fengqi and You, Zebin and Zhang, Xiaolu and Ou, Jingyang and Hu, Jun and Zhou, Jun and Lin, Yankai and Wen, Ji-Rong and Li, Chongxuan},
  journal = {arXiv preprint arXiv:2502.09992},
  year    = {2025}
}

@inproceedings{arriola2025blockdiffusion,
  title     = {Block diffusion: Interpolating between autoregressive and diffusion language models},
  author    = {Arriola, Marianne and Gokaslan, Aaron and Chiu, Justin and Yang, Zhihan and Qi, Zhixuan and Han, Jiaqi and Sahoo, Subham and Kuleshov, Volodymyr},
  booktitle = {International Conference on Learning Representations},
  volume    = {2025},
  pages     = {50726--50753},
  year      = {2025}
}

@article{assran2025vjepa2,
  title   = {V-jepa 2: Self-supervised video models enable understanding, prediction and planning},
  author  = {Assran, Mido and Bardes, Adrien and Fan, David and Garrido, Quentin and Howes, Russell and Muckley, Matthew and Rizvi, Ammar and Roberts, Claire and Sinha, Koustuv and Zholus, Artem and others},
  journal = {arXiv preprint arXiv:2506.09985},
  year    = {2025}
}

@inproceedings{bahmani2024tc4d,
  title         = {TC4D: Trajectory-Conditioned Text-to-4D Generation},
  author        = {Sherwin Bahmani and Xian Liu and Wang Yifan and Ivan Skorokhodov and Victor Rong and Ziwei Liu and Xihui Liu and Jeong Joon Park and Sergey Tulyakov and Gordon Wetzstein and Andrea Tagliasacchi and David B. Lindell},
  booktitle     = {Proceedings of the European Conference on Computer Vision (ECCV)},
  year          = {2024},
  eprint        = {2403.17920},
  archiveprefix = {arXiv}
}

@inproceedings{barquero2024flowmdm,
  title         = {Seamless Human Motion Composition with Blended Positional Encodings},
  author        = {German Barquero and Sergio Escalera and Cristina Palmero},
  booktitle     = {Proceedings of the IEEE/CVF Conference on Computer Vision and Pattern Recognition (CVPR)},
  year          = {2024},
  eprint        = {2402.15509},
  archiveprefix = {arXiv}
}

@article{black2024pi0,
  title   = {$\pi_0$: A Vision-Language-Action Flow Model for General Robot Control},
  author  = {Black, Kevin and Brown, Noah and Driess, Danny and Esmail, Adnan and Equi, Michael and Finn, Chelsea and Fusai, Niccolo and Groom, Lachy and Hausman, Karol and Ichter, Brian and others},
  journal = {arXiv preprint arXiv:2410.24164},
  year    = {2024}
}

@article{blattmann2023svd,
  title         = {Stable Video Diffusion: Scaling Latent Video Diffusion Models to Large Datasets},
  author        = {Blattmann, Andreas and Dockhorn, Tim and Kulal, Sumith and Mendelevitch, Daniel and Kilian, Maciej and Lorenz, Dominik and Levi, Yam and English, Zion and Voleti, Vikram and Letts, Adam and others},
  journal       = {arXiv preprint arXiv:2311.15127},
  year          = {2023},
  eprint        = {2311.15127},
  archiveprefix = {arXiv}
}

@inproceedings{cai2026flooddiffusion,
  title     = {Flooddiffusion: Tailored diffusion forcing for streaming motion generation},
  author    = {Cai, Yiyi and Wu, Yuhan and Li, Kunhang and Zhou, You and Zheng, Bo and Liu, Haiyang},
  booktitle = {Proceedings of the IEEE/CVF Conference on Computer Vision and Pattern Recognition},
  pages     = {2295--2304},
  year      = {2026}
}

@inproceedings{chang2022maskgit,
  title     = {{MaskGIT}: Masked Generative Image Transformer},
  author    = {Chang, Huiwen and Zhang, Han and Jiang, Lu and Liu, Ce and Freeman, William T.},
  booktitle = {IEEE/CVF Conference on Computer Vision and Pattern Recognition (CVPR)},
  year      = {2022}
}

@inproceedings{chen2024diffusionforcing,
  title         = {Diffusion Forcing: Next-token Prediction Meets Full-Sequence Diffusion},
  author        = {Boyuan Chen and Diego Marti Monso and Yilun Du and Max Simchowitz and Russ Tedrake and Vincent Sitzmann},
  booktitle     = {Advances in Neural Information Processing Systems (NeurIPS)},
  year          = {2024},
  eprint        = {2407.01392},
  archiveprefix = {arXiv}
}

@inproceedings{chen2024moto,
  title         = {Moto: Latent Motion Token as the Bridging Language for Learning Robot Manipulation from Videos},
  author        = {Yi Chen and Yuying Ge and Weiliang Tang and Yizhuo Li and Yixiao Ge and Mingyu Ding and Ying Shan and Xihui Liu},
  booktitle     = {Proceedings of the IEEE/CVF International Conference on Computer Vision (ICCV)},
  year          = {2025},
  eprint        = {2412.04445},
  archiveprefix = {arXiv}
}

@inproceedings{chen2025ecflow,
  title         = {EC-Flow: Enabling Versatile Robotic Manipulation from Action-Unlabeled Videos via Embodiment-Centric Flow},
  author        = {Chen, Yixiang and Li, Peiyan and Huang, Yan and Yang, Jiabing and Chen, Kehan and Wang, Liang},
  booktitle     = {Proceedings of the IEEE/CVF International Conference on Computer Vision (ICCV)},
  year          = {2025},
  eprint        = {2507.06224},
  archiveprefix = {arXiv}
}

@article{chi2023diffusionpolicy,
  title     = {Diffusion policy: Visuomotor policy learning via action diffusion},
  author    = {Chi, Cheng and Xu, Zhenjia and Feng, Siyuan and Cousineau, Eric and Du, Yilun and Burchfiel, Benjamin and Tedrake, Russ and Song, Shuran},
  journal   = {The International Journal of Robotics Research},
  volume    = {44},
  number    = {10-11},
  pages     = {1684--1704},
  year      = {2025},
  publisher = {Sage Publications Sage UK: London, England}
}

@misc{cong2026umo,
  title         = {UMO: Unified In-Context Learning Unlocks Motion Foundation Model Priors},
  author        = {Xiaoyan Cong and Zekun Li and Zhiyang Dou and Hongyu Li and Omid Taheri and Chuan Guo and Abhay Mittal and Sizhe An and Taku Komura and Wojciech Matusik and Michael J. Black and Srinath Sridhar},
  year          = {2026},
  eprint        = {2603.15975},
  archiveprefix = {arXiv}
}

@article{deng2024causaldit,
  title   = {Causal diffusion transformers for generative modeling},
  author  = {Deng, Chaorui and Zhu, Deyao and Li, Kunchang and Guang, Shi and Fan, Haoqi},
  journal = {arXiv preprint arXiv:2412.12095},
  year    = {2024}
}

@inproceedings{diller2024cghoi,
  title         = {CG-HOI: Contact-Guided 3D Human-Object Interaction Generation},
  author        = {Diller, Christian and Dai, Angela},
  booktitle     = {CVPR},
  year          = {2024},
  eprint        = {2311.16097},
  archiveprefix = {arXiv}
}

@inproceedings{du2023unipi,
  title     = {Learning Universal Policies via Text-Guided Video Generation},
  author    = {Du, Yilun and Yang, Mengjiao and Dai, Bo and Dai, Hanjun and Nachum, Ofir and Tenenbaum, Joshua B. and Schuurmans, Dale and Abbeel, Pieter},
  booktitle = {Advances in Neural Information Processing Systems (NeurIPS)},
  year      = {2023}
}

@misc{duran2026fusion,
  title         = {FUSION: Full-Body Unified Motion Prior for Body and Hands via Diffusion},
  author        = {Duran, Enes and Athanasiou, Nikos and Kocabas, Muhammed and Black, Michael J. and Taheri, Omid},
  year          = {2026},
  eprint        = {2601.03959},
  archiveprefix = {arXiv}
}

@article{gao2026dreamdojo,
  title   = {DreamDojo: A Generalist Robot World Model from Large-Scale Human Videos},
  author  = {Gao, Shenyuan and Liang, William and Zheng, Kaiyuan and Malik, Ayaan and Ye, Seonghyeon and Yu, Sihyun and Tseng, Wei-Cheng and Dong, Yuzhu and Mo, Kaichun and Lin, Chen-Hsuan and Ma, Qianli and Nah, Seungjun and Magne, Loic and Xiang, Jiannan and Xie, Yuqi and Zheng, Ruijie and Niu, Dantong and Tan, You Liang and Zentner, K. R. and Kurian, George and Indupuru, Suneel and Jannaty, Pooya and Gu, Jinwei and Zhang, Jun and Malik, Jitendra and Abbeel, Pieter and Liu, Ming-Yu and Zhu, Yuke and Jang, Joel and Fan, Linxi},
  journal = {arXiv preprint arXiv:2602.06949},
  year    = {2026}
}

@inproceedings{goel2023humans4d,
  title         = {Humans in 4D: Reconstructing and Tracking Humans with Transformers},
  author        = {Shubham Goel and Georgios Pavlakos and Jathushan Rajasegaran and Angjoo Kanazawa and Jitendra Malik},
  booktitle     = {Proceedings of the IEEE/CVF International Conference on Computer Vision (ICCV)},
  year          = {2023},
  eprint        = {2305.20091},
  archiveprefix = {arXiv}
}

@article{hafner2023dreamerv3,
  title         = {Mastering diverse control tasks through world models},
  author        = {Danijar Hafner and Jurgis Pasukonis and Jimmy Ba and Timothy Lillicrap},
  journal       = {Nature},
  year          = {2025},
  eprint        = {2301.04104},
  archiveprefix = {arXiv}
}

@inproceedings{hansen2024hwm,
  title     = {Hierarchical world models as visual whole-body humanoid controllers},
  author    = {Hansen, Nick and SV, Jyothir and Sobal, Vlad and LeCun, Yann and Wang, Xiaolong and Su, Hao},
  booktitle = {International Conference on Learning Representations},
  volume    = {2025},
  pages     = {62175--62195},
  year      = {2025}
}

@article{he2025crossembodimentwm,
  title   = {Scaling Cross-Embodiment World Models for Dexterous Manipulation},
  author  = {He, Zihao and Ai, Bo and Mu, Tongzhou and Liu, Yulin and Wan, Weikang and Fu, Jiawei and Du, Yilun and Christensen, Henrik I. and Su, Hao},
  journal = {arXiv preprint arXiv:2511.01177},
  year    = {2025}
}

@article{ho2022cfg,
  title         = {Classifier-Free Diffusion Guidance},
  author        = {Ho, Jonathan and Salimans, Tim},
  journal       = {arXiv preprint arXiv:2207.12598},
  year          = {2022},
  eprint        = {2207.12598},
  archiveprefix = {arXiv},
  note          = {Short version in NeurIPS 2021 Workshop on Deep Generative Models and Downstream Applications}
}

@inproceedings{hoeg2024streamingdp,
  title         = {Streaming Diffusion Policy: Fast Policy Synthesis with Variable Noise Diffusion Models},
  author        = {Sigmund H. Hoeg and Yilun Du and Olav Egeland},
  booktitle     = {IEEE International Conference on Robotics and Automation (ICRA)},
  year          = {2025},
  eprint        = {2406.04806},
  archiveprefix = {arXiv}
}

@inproceedings{hsu2024spot,
  title     = {SPOT: SE(3) Pose Trajectory Diffusion for Object-Centric Manipulation},
  author    = {Hsu, Cheng-Chun and Wen, Bowen and Xu, Jie and Narang, Yashraj and Wang, Xiaolong and Zhu, Yuke and Biswas, Joydeep and Birchfield, Stan},
  booktitle = {IEEE International Conference on Robotics and Automation (ICRA)},
  year      = {2025}
}

@article{huang2024multimodaldf,
  title   = {Unified Multimodal Diffusion Forcing for Forceful Manipulation},
  author  = {Huang, Zixuan and Hou, Huaidian and Berenson, Dmitry},
  journal = {arXiv preprint arXiv:2511.04812},
  year    = {2025}
}

@article{huang2025memoryforcing,
  title         = {Memory Forcing: Spatio-Temporal Memory for Consistent Scene Generation on Minecraft},
  author        = {Junchao Huang and Xinting Hu and Boyao Han and Shaoshuai Shi and Zhuotao Tian and Tianyu He and Li Jiang},
  journal       = {arXiv preprint arXiv:2510.03198},
  year          = {2025},
  eprint        = {2510.03198},
  archiveprefix = {arXiv}
}

@article{huang2025selfforcing,
  title   = {Self forcing: Bridging the train-test gap in autoregressive video diffusion},
  author  = {Huang, Xun and Li, Zhengqi and He, Guande and Zhou, Mingyuan and Shechtman, Eli},
  journal = {Advances in Neural Information Processing Systems},
  volume  = {38},
  pages   = {167283--167308},
  year    = {2026}
}

@misc{huang2026pointworld,
  title         = {PointWorld: Scaling 3D World Models for In-The-Wild Robotic Manipulation},
  author        = {Huang, Wenlong and Chao, Yu-Wei and Mousavian, Arsalan and Liu, Ming-Yu and Fox, Dieter and Mo, Kaichun and Fei-Fei, Li},
  year          = {2026},
  eprint        = {2601.03782},
  archiveprefix = {arXiv}
}

@inproceedings{janner2022diffuser,
  title         = {Planning with Diffusion for Flexible Behavior Synthesis},
  author        = {Michael Janner and Yilun Du and Joshua B. Tenenbaum and Sergey Levine},
  booktitle     = {International Conference on Machine Learning (ICML)},
  year          = {2022},
  eprint        = {2205.09991},
  archiveprefix = {arXiv}
}

@inproceedings{jiang2024motiongpt,
  title         = {MotionGPT: Human Motion as a Foreign Language},
  author        = {Biao Jiang and Xin Chen and Wen Liu and Jingyi Yu and Gang Yu and Tao Chen},
  booktitle     = {Advances in Neural Information Processing Systems (NeurIPS)},
  year          = {2023},
  eprint        = {2306.14795},
  archiveprefix = {arXiv}
}

@misc{lee2025tracegen,
  title         = {TraceGen: World Modeling in 3D Trace Space Enables Learning from Cross-Embodiment Videos},
  author        = {Lee, Seungjae and Jung, Yoonkyo and Chun, Inkook and Lee, Yao-Chih and Cai, Zikui and Huang, Hongjia and Talreja, Aayush and Dao, Tan Dat and Liang, Yongyuan and Huang, Jia-Bin and Huang, Furong},
  year          = {2025},
  eprint        = {2511.21690},
  archiveprefix = {arXiv}
}

@inproceedings{lei2024mosca,
  title         = {MoSca: Dynamic Gaussian Fusion from Casual Videos via 4D Motion Scaffolds},
  author        = {Jiahui Lei and Yijia Weng and Adam Harley and Leonidas Guibas and Kostas Daniilidis},
  booktitle     = {Proceedings of the IEEE/CVF Conference on Computer Vision and Pattern Recognition (CVPR)},
  year          = {2025},
  eprint        = {2405.17421},
  archiveprefix = {arXiv}
}

@inproceedings{lei2025momaps,
  title         = {MoMaps: Semantics-Aware Scene Motion Generation with Motion Maps},
  author        = {Lei, Jiahui and Genova, Kyle and Kopanas, George and Snavely, Noah and Guibas, Leonidas},
  booktitle     = {Proceedings of the IEEE/CVF International Conference on Computer Vision (ICCV)},
  year          = {2025},
  eprint        = {2510.11107},
  archiveprefix = {arXiv}
}

@article{li2023omomo,
  title         = {Object Motion Guided Human Motion Synthesis},
  author        = {Jiaman Li and Jiajun Wu and C. Karen Liu},
  journal       = {ACM Transactions on Graphics (Proc. SIGGRAPH Asia)},
  year          = {2023},
  eprint        = {2309.16237},
  archiveprefix = {arXiv}
}

@inproceedings{li2024chois,
  title         = {Controllable Human-Object Interaction Synthesis},
  author        = {Li, Jiaman and Clegg, Alexander and Mottaghi, Roozbeh and Wu, Jiajun and Puig, Xavier and Liu, C. Karen},
  booktitle     = {European Conference on Computer Vision (ECCV)},
  year          = {2024},
  eprint        = {2312.03913},
  archiveprefix = {arXiv}
}

@inproceedings{li2025uva,
  title         = {Unified Video Action Model},
  author        = {Shuang Li and Yihuai Gao and Dorsa Sadigh and Shuran Song},
  booktitle     = {Robotics: Science and Systems (RSS)},
  year          = {2025},
  eprint        = {2503.00200},
  archiveprefix = {arXiv}
}

@article{li2026llamomotion,
  title         = {LLaMo: Scaling Pretrained Language Models for Unified Motion Understanding and Generation with Continuous Autoregressive Tokens},
  author        = {Zekun Li and Sizhe An and Chengcheng Tang and Chuan Guo and Ivan Shugurov and Linguang Zhang and Amy Zhao and Srinath Sridhar and Lingling Tao and Abhay Mittal},
  journal       = {arXiv preprint arXiv:2602.12370},
  year          = {2026},
  eprint        = {2602.12370},
  archiveprefix = {arXiv}
}

@inproceedings{lipman2022flowmatching,
  title         = {Flow Matching for Generative Modeling},
  author        = {Yaron Lipman and Ricky T. Q. Chen and Heli Ben-Hamu and Maximilian Nickel and Matt Le},
  booktitle     = {International Conference on Learning Representations (ICLR)},
  year          = {2023},
  eprint        = {2210.02747},
  archiveprefix = {arXiv}
}

@inproceedings{liu2022rectified,
  title         = {Flow Straight and Fast: Learning to Generate and Transfer Data with Rectified Flow},
  author        = {Xingchao Liu and Chengyue Gong and Qiang Liu},
  booktitle     = {International Conference on Learning Representations (ICLR)},
  year          = {2023},
  eprint        = {2209.03003},
  archiveprefix = {arXiv}
}

@article{liu2025geom4drobot,
  title   = {Geometry-aware 4d video generation for robot manipulation},
  author  = {Liu, Zeyi and Li, Shuang and Cousineau, Eric and Feng, Siyuan and Burchfiel, Benjamin and Song, Shuran},
  journal = {arXiv preprint arXiv:2507.01099},
  year    = {2025}
}

@misc{liu2025rollingforcing,
  title         = {Rolling Forcing: Autoregressive Long Video Diffusion in Real Time},
  author        = {Kunhao Liu and Wenbo Hu and Jiale Xu and Ying Shan and Shijian Lu},
  year          = {2025},
  eprint        = {2509.25161},
  archiveprefix = {arXiv}
}

@article{liu2026worldvla,
  title         = {World-VLA-Loop: Closed-Loop Learning of Video World Model and VLA Policy},
  author        = {Xiaokang Liu and Zechen Bai and Hai Ci and Kevin Yuchen Ma and Mike Zheng Shou},
  journal       = {arXiv preprint arXiv:2602.06508},
  year          = {2026},
  eprint        = {2602.06508},
  archiveprefix = {arXiv}
}

@inproceedings{lugmayr2022repaint,
  title     = {Repaint: Inpainting using denoising diffusion probabilistic models},
  author    = {Lugmayr, Andreas and Danelljan, Martin and Romero, Andres and Yu, Fisher and Timofte, Radu and Van Gool, Luc},
  booktitle = {Proceedings of the IEEE/CVF conference on computer vision and pattern recognition},
  pages     = {11461--11471},
  year      = {2022}
}

@misc{maluleke2025magnet,
  title         = {Diffusion Forcing for Multi-Agent Interaction Sequence Modeling},
  author        = {Maluleke, Vongani H. and Horiuchi, Kie and Wilken, Lea and Ng, Evonne and Malik, Jitendra and Kanazawa, Angjoo},
  year          = {2025},
  eprint        = {2512.17900},
  archiveprefix = {arXiv}
}

@article{nvidia2025cosmos,
  title   = {Cosmos World Foundation Model Platform for Physical AI},
  author  = {{NVIDIA} and Agarwal, Niket and Ali, Arslan and Bala, Maciej and Balaji, Yogesh and Barker, Erik and Cai, Tiffany and Chattopadhyay, Prithvijit and Chen, Yongxin and Cui, Yin and Ding, Yifan and Dworakowski, Daniel and Fan, Jiaojiao and Fenzi, Michele and Ferroni, Francesco and Fidler, Sanja and Fox, Dieter and Ge, Songwei and others},
  journal = {arXiv preprint arXiv:2501.03575},
  year    = {2025}
}

@inproceedings{peebles2023dit,
  title         = {Scalable Diffusion Models with Transformers},
  author        = {William Peebles and Saining Xie},
  booktitle     = {Proceedings of the IEEE/CVF International Conference on Computer Vision (ICCV)},
  year          = {2023},
  eprint        = {2212.09748},
  archiveprefix = {arXiv}
}

@article{reed2022gato,
  title   = {A Generalist Agent},
  author  = {Reed, Scott and Zolna, Konrad and Parisotto, Emilio and Colmenarejo, Sergio Gomez and Novikov, Alexander and Barth-Maron, Gabriel and Gimenez, Mai and Sulsky, Yury and Kay, Jackie and Springenberg, Jost Tobias and Eccles, Tom and Bruce, Jake and Razavi, Ali and Edwards, Ashley and Heess, Nicolas and Chen, Yutian and Hadsell, Raia and Vinyals, Oriol and Bordbar, Mahyar and de Freitas, Nando},
  journal = {Transactions on Machine Learning Research},
  year    = {2022}
}

@article{tessler2024maskedmimic,
  title         = {MaskedMimic: Unified Physics-Based Character Control Through Masked Motion Inpainting},
  author        = {Chen Tessler and Yunrong Guo and Ofir Nabati and Gal Chechik and Xue Bin Peng},
  journal       = {ACM Transactions on Graphics},
  year          = {2024},
  eprint        = {2409.14393},
  archiveprefix = {arXiv}
}

@article{tevet2022mdm,
  title   = {Human motion diffusion model},
  author  = {Tevet, Guy and Raab, Sigal and Gordon, Brian and Shafir, Yonatan and Cohen-Or, Daniel and Bermano, Amit H},
  journal = {arXiv preprint arXiv:2209.14916},
  year    = {2022}
}

@misc{wang2025discretedf,
  title         = {Diffusion LLMs Can Do Faster-Than-AR Inference via Discrete Diffusion Forcing},
  author        = {Xu Wang and Chenkai Xu and Yijie Jin and Jiachun Jin and Hao Zhang and Zhijie Deng},
  year          = {2025},
  eprint        = {2508.09192},
  archiveprefix = {arXiv}
}

@inproceedings{wen2024atm,
  title         = {Any-point Trajectory Modeling for Policy Learning},
  author        = {Wen, Chuan and Lin, Xingyu and So, John and Chen, Kai and Dou, Qi and Gao, Yang and Abbeel, Pieter},
  booktitle     = {Robotics: Science and Systems (RSS)},
  year          = {2024},
  eprint        = {2401.00025},
  archiveprefix = {arXiv}
}

@inproceedings{xu2023interdiff,
  title         = {InterDiff: Generating 3D Human-Object Interactions with Physics-Informed Diffusion},
  author        = {Sirui Xu and Zhengyuan Li and Yu-Xiong Wang and Liang-Yan Gui},
  booktitle     = {Proceedings of the IEEE/CVF International Conference on Computer Vision (ICCV)},
  year          = {2023},
  eprint        = {2308.16905},
  archiveprefix = {arXiv}
}

@article{xu2024comp4d,
  title   = {Comp4d: Llm-guided compositional 4d scene generation},
  author  = {Xu, Dejia and Liang, Hanwen and Bhatt, Neel P and Hu, Hezhen and Liang, Hanxue and Plataniotis, Konstantinos N and Wang, Zhangyang},
  journal = {arXiv preprint arXiv:2403.16993},
  year    = {2024}
}

@inproceedings{yang2024cogvideox,
  title         = {CogVideoX: Text-to-Video Diffusion Models with An Expert Transformer},
  author        = {Yang, Zhuoyi and Teng, Jiayan and Zheng, Wendi and Ding, Ming and Huang, Shiyu and Xu, Jiazheng and Yang, Yuanming and Hong, Wenyi and Zhang, Xiaohan and Feng, Guanyu and others},
  booktitle     = {International Conference on Learning Representations (ICLR)},
  year          = {2025},
  eprint        = {2408.06072},
  archiveprefix = {arXiv}
}

@misc{ye2026dreamzero,
  title         = {World Action Models are Zero-shot Policies},
  author        = {Ye, Seonghyeon and Ge, Yunhao and Zheng, Kaiyuan and Gao, Shenyuan and Yu, Sihyun and Kurian, George and Indupuru, Suneel and Tan, You Liang and Zhu, Chuning and Xiang, Jiannan and Malik, Ayaan and Lee, Kyungmin and Liang, William and Ranawaka, Nadun and Gu, Jiasheng and Xu, Yinzhen and Wang, Guanzhi and Hu, Fengyuan and Narayan, Avnish and Bjorck, Johan and Wang, Jing and Kim, Gwanghyun and Niu, Dantong and Zheng, Ruijie and Xie, Yuqi and Wu, Jimmy and Wang, Qi and Julian, Ryan and Xu, Danfei and Du, Yilun and Chebotar, Yevgen and Reed, Scott and Kautz, Jan and Zhu, Yuke and Fan, Linxi and Jang, Joel},
  year          = {2026},
  eprint        = {2602.15922},
  archiveprefix = {arXiv}
}

@article{yuan2024generalflow,
  title   = {General flow as foundation affordance for scalable robot learning},
  author  = {Yuan, Chengbo and Wen, Chuan and Zhang, Tong and Gao, Yang},
  journal = {arXiv preprint arXiv:2401.11439},
  year    = {2024}
}

@inproceedings{zhao2023act,
  title         = {Learning Fine-Grained Bimanual Manipulation with Low-Cost Hardware},
  author        = {Tony Z. Zhao and Vikash Kumar and Sergey Levine and Chelsea Finn},
  booktitle     = {Robotics: Science and Systems (RSS)},
  year          = {2023},
  eprint        = {2304.13705},
  archiveprefix = {arXiv}
}

@inproceedings{zhen2024threedvla,
  title         = {3D-VLA: A 3D Vision-Language-Action Generative World Model},
  author        = {Haoyu Zhen and Xiaowen Qiu and Peihao Chen and Jincheng Yang and Xin Yan and Yilun Du and Yining Hong and Chuang Gan},
  booktitle     = {International Conference on Machine Learning (ICML)},
  year          = {2024},
  eprint        = {2403.09631},
  archiveprefix = {arXiv}
}

@inproceedings{zholus2025tapnext,
  title         = {TAPNext: Tracking Any Point ({TAP}) as Next Token Prediction},
  author        = {Zholus, Artem and Doersch, Carl and others},
  booktitle     = {Proceedings of the IEEE/CVF International Conference on Computer Vision (ICCV)},
  year          = {2025},
  eprint        = {2504.05579},
  archiveprefix = {arXiv}
}

@inproceedings{zhu2025uwm,
  title     = {Unified World Models: Coupling Video and Action Diffusion for Pretraining on Large Robotic Datasets},
  author    = {Zhu, Chuning and Yu, Raymond and Feng, Siyuan and Burchfiel, Benjamin and Shah, Paarth and Gupta, Abhishek},
  booktitle = {Robotics: Science and Systems (RSS)},
  year      = {2025}
}

@inproceedings{zhang2024largemotionmodel,
  title         = {Large Motion Model for Unified Multi-Modal Motion Generation},
  author        = {Mingyuan Zhang and Daisheng Jin and Chenyang Gu and Fangzhou Hong and Zhongang Cai and Jingfang Huang and Chongzhi Zhang and Xinying Guo and Lei Yang and Ying He and Ziwei Liu},
  booktitle     = {Proceedings of the European Conference on Computer Vision (ECCV)},
  year          = {2024},
  eprint        = {2404.01284},
  archiveprefix = {arXiv}
}

@misc{dex4d2026,
  title         = {Dex4D: Task-Agnostic Point Track Policy for Sim-to-Real Dexterous Manipulation},
  author        = {Yuxuan Kuang and Sungjae Park and Katerina Fragkiadaki and Shubham Tulsiani},
  year          = {2026},
  eprint        = {2602.15828},
  archiveprefix = {arXiv},
  note          = {arXiv:2602.15828}
}

@misc{castillo2023bodiffusion,
  title         = {BoDiffusion: Diffusing Sparse Observations for Full-Body Human Motion Synthesis},
  author        = {Angela Castillo and Maria Escobar and Guillaume Jeanneret and Albert Pumarola and Pablo Arbelaez and Ali Thabet and Artsiom Sanakoyeu},
  year          = {2023},
  eprint        = {2304.11118},
  archiveprefix = {arXiv},
  note          = {arXiv:2304.11118}
}

@inproceedings{wang2025scalinglmm,
  title         = {Scaling Large Motion Models with Million-Level Human Motions},
  author        = {Ye Wang and Sipeng Zheng and Bin Cao and Qianshan Wei and Weishuai Zeng and Qin Jin and Zongqing Lu},
  booktitle     = {International Conference on Machine Learning (ICML)},
  year          = {2025},
  eprint        = {2410.03311},
  archiveprefix = {arXiv}
}

@misc{baharani2025mofm,
  title         = {MoFM: A Large-Scale Human Motion Foundation Model},
  author        = {Mohammadreza Baharani and Ghazal Alinezhad Noghre and Armin Danesh Pazho and Gabriel Maldonado and Hamed Tabkhi},
  year          = {2025},
  eprint        = {2502.05432},
  archiveprefix = {arXiv},
  note          = {arXiv:2502.05432}
}

@inproceedings{patel2025uniegomotion,
  title         = {UniEgoMotion: A Unified Model for Egocentric Motion Reconstruction, Forecasting, and Generation},
  author        = {Chaitanya Patel and Hiroki Nakamura and Yuta Kyuragi and Kazuki Kozuka and Juan Carlos Niebles and Ehsan Adeli},
  booktitle     = {Proceedings of the IEEE/CVF International Conference on Computer Vision (ICCV)},
  year          = {2025},
  eprint        = {2508.01126},
  archiveprefix = {arXiv}
}

@inproceedings{xu2025multiperson,
  title         = {Multi-Person Interaction Generation from Two-Person Motion Priors},
  author        = {Wenning Xu and Shiyu Fan and Paul Henderson and Edmond S. L. Ho},
  booktitle     = {ACM SIGGRAPH Conference Papers},
  year          = {2025},
  eprint        = {2505.17860},
  archiveprefix = {arXiv}
}

@misc{forge2025difftree,
  title         = {FORGE-Tree: Diffusion-Forcing Tree Search for Long-Horizon Robot Manipulation},
  author        = {Yanjia Huang and Shuo Liu and Sheng Liu and Qingxiao Xu and Mingyang Wu and Xiangbo Gao and Zhengzhong Tu},
  year          = {2025},
  eprint        = {2510.21744},
  archiveprefix = {arXiv},
  note          = {arXiv:2510.21744}
}

@inproceedings{lyu2025dywa,
  title         = {DyWA: Dynamics-Adaptive World Action Model for Generalizable Non-Prehensile Manipulation},
  author        = {Jiangran Lyu and Ziming Li and Xuesong Shi and Chaoyi Xu and Yizhou Wang and He Wang},
  booktitle     = {Proceedings of the IEEE/CVF International Conference on Computer Vision (ICCV)},
  year          = {2025},
  eprint        = {2503.16806},
  archiveprefix = {arXiv}
}

@misc{zhan2026perpetualwonder,
  title         = {PerpetualWonder: Long-Horizon Action-Conditioned 4D Scene Generation},
  author        = {Jiahao Zhan and Zizhang Li and Hong-Xing Yu and Jiajun Wu},
  year          = {2026},
  eprint        = {2602.04876},
  archiveprefix = {arXiv},
  note          = {arXiv:2602.04876}
}

@article{mu2025smp,
  title   = {Smp: Reusable score-matching motion priors for physics-based character control},
  author  = {Mu, Yuxuan and Zhang, Ziyu and Shi, Yi and Yang, Dun and Matsumoto, Minami and Imamura, Kotaro and Tevet, Guy and Guo, Chuan and Taylor, Michael and Shu, Chang and others},
  journal = {arXiv preprint arXiv:2512.03028},
  year    = {2025}
}

@article{huang2024imaginationpolicy,
  title   = {Imagination policy: Using generative point cloud models for learning manipulation policies},
  author  = {Huang, Haojie and Schmeckpeper, Karl and Wang, Dian and Biza, Ondrej and Qian, Yaoyao and Liu, Haotian and Jia, Mingxi and Platt, Robert and Walters, Robin},
  journal = {arXiv preprint arXiv:2406.11740},
  year    = {2024}
}

@article{autoworld2026,
  title   = {AutoWorld: Scaling Multi-Agent Traffic Simulation with Self-Supervised World Models},
  author  = {Pourkeshavatz, Mozhgan and Liu, Tianran and Rhinehart, Nicholas},
  journal = {arXiv preprint arXiv:2603.28963},
  year    = {2026}
}

@inproceedings{zhou2024dinowm,
  title         = {DINO-WM: World Models on Pre-trained Visual Features enable Zero-shot Planning},
  author        = {Gaoyue Zhou and Hengkai Pan and Yann LeCun and Lerrel Pinto},
  booktitle     = {International Conference on Machine Learning (ICML)},
  year          = {2025},
  eprint        = {2411.04983},
  archiveprefix = {arXiv}
}

@inproceedings{byravan2017se3nets,
  title         = {{SE3-Nets}: Learning Rigid Body Motion using Deep Neural Networks},
  author        = {Byravan, Arunkumar and Fox, Dieter},
  booktitle     = {IEEE International Conference on Robotics and Automation (ICRA)},
  year          = {2017},
  eprint        = {1606.02378},
  archiveprefix = {arXiv}
}

@article{byravan2018se3posenets,
  title   = {Se3-pose-nets: Structured deep dynamics models for visuomotor planning and control},
  author  = {Byravan, Arunkumar and Leeb, Felix and Meier, Franziska and Fox, Dieter},
  journal = {arXiv preprint arXiv:1710.00489},
  year    = {2017}
}

@inproceedings{kratimenos2024dynmf,
  title         = {{DynMF}: Neural Motion Factorization for Real-time Dynamic View Synthesis with {3D} Gaussian Splatting},
  author        = {Agelos Kratimenos and Jiahui Lei and Kostas Daniilidis},
  booktitle     = {Proceedings of the European Conference on Computer Vision (ECCV)},
  year          = {2024},
  eprint        = {2312.00112},
  archiveprefix = {arXiv}
}

@inproceedings{wang2024som,
  title     = {Shape of motion: 4d reconstruction from a single video},
  author    = {Wang, Qianqian and Ye, Vickie and Gao, Hang and Zeng, Weijia and Austin, Jake and Li, Zhengqi and Kanazawa, Angjoo},
  booktitle = {Proceedings of the IEEE/CVF International Conference on Computer Vision},
  pages     = {9660--9672},
  year      = {2025}
}

@article{oquab2023dinov2,
  title         = {{DINO}v2: Learning Robust Visual Features without Supervision},
  author        = {Maxime Oquab and Timothée Darcet and Théo Moutakanni and Huy V. Vo and Marc Szafraniec and Vasil Khalidov and Pierre Fernandez and Daniel Haziza and Francisco Massa and Alaaeldin El-Nouby and Mahmoud Assran and Nicolas Ballas and Wojciech Galuba and Russell Howes and Po-Yao Huang and Shang-Wen Li and Ishan Misra and Michael Rabbat and Vasu Sharma and Gabriel Synnaeve and Hu Xu and Hervé Jegou and Julien Mairal and Patrick Labatut and Armand Joulin and Piotr Bojanowski},
  journal       = {Transactions on Machine Learning Research},
  year          = {2024},
  eprint        = {2304.07193},
  archiveprefix = {arXiv}
}

@inproceedings{zhou2019continuity,
  title         = {On the Continuity of Rotation Representations in Neural Networks},
  author        = {Zhou, Yi and Barnes, Connelly and Lu, Jingwan and Yang, Jimei and Li, Hao},
  booktitle     = {IEEE Conference on Computer Vision and Pattern Recognition (CVPR)},
  year          = {2019},
  eprint        = {1812.07035},
  archiveprefix = {arXiv}
}

@article{loper2015smpl,
  title   = {{SMPL}: A Skinned Multi-Person Linear Model},
  author  = {Loper, Matthew and Mahmood, Naureen and Romero, Javier and Pons-Moll, Gerard and Black, Michael J.},
  journal = {ACM Transactions on Graphics (Proc.\ SIGGRAPH Asia)},
  volume  = {34},
  number  = {6},
  pages   = {248:1--248:16},
  year    = {2015}
}

@article{romero2017mano,
  title   = {Embodied Hands: Modeling and Capturing Hands and Bodies Together},
  author  = {Romero, Javier and Tzionas, Dimitrios and Black, Michael J.},
  journal = {ACM Transactions on Graphics (Proc.\ SIGGRAPH Asia)},
  volume  = {36},
  number  = {6},
  pages   = {245:1--245:17},
  year    = {2017}
}

@inproceedings{zuffi2017smal,
  title     = {{3D} Menagerie: Modeling the {3D} Shape and Pose of Animals},
  author    = {Zuffi, Silvia and Kanazawa, Angjoo and Jacobs, David W. and Black, Michael J.},
  booktitle = {IEEE Conference on Computer Vision and Pattern Recognition (CVPR)},
  year      = {2017}
}

@inproceedings{lei2024gart,
  title         = {{GART}: Gaussian Articulated Template Models},
  author        = {Jiahui Lei and Yufu Wang and Georgios Pavlakos and Lingjie Liu and Kostas Daniilidis},
  booktitle     = {Proceedings of the IEEE/CVF Conference on Computer Vision and Pattern Recognition (CVPR)},
  year          = {2024},
  eprint        = {2311.16099},
  archiveprefix = {arXiv}
}

@inproceedings{pavlakos2024hamer,
  title         = {Reconstructing Hands in {3D} with Transformers},
  author        = {Georgios Pavlakos and Dandan Shan and Ilija Radosavovic and Angjoo Kanazawa and David Fouhey and Jitendra Malik},
  booktitle     = {Proceedings of the IEEE/CVF Conference on Computer Vision and Pattern Recognition (CVPR)},
  year          = {2024},
  eprint        = {2312.05251},
  archiveprefix = {arXiv}
}

@inproceedings{darcet2024registers,
  title         = {Vision Transformers Need Registers},
  author        = {Darcet, Timoth{\'e}e and Oquab, Maxime and Mairal, Julien and Bojanowski, Piotr},
  booktitle     = {International Conference on Learning Representations (ICLR)},
  year          = {2024},
  eprint        = {2309.16588},
  archiveprefix = {arXiv}
}

@article{lynch2022langtable,
  title     = {Interactive language: Talking to robots in real time},
  author    = {Lynch, Corey and Wahid, Ayzaan and Tompson, Jonathan and Ding, Tianli and Betker, James and Baruch, Robert and Armstrong, Travis and Florence, Pete},
  journal   = {IEEE Robotics and Automation Letters},
  year      = {2023},
  publisher = {IEEE}
}

@inproceedings{ye2024lapa,
  title         = {Latent Action Pretraining from Videos},
  author        = {Seonghyeon Ye and Joel Jang and Byeongguk Jeon and Sejune Joo and Jianwei Yang and Baolin Peng and Ajay Mandlekar and Reuben Tan and Yu-Wei Chao and Bill Yuchen Lin and Lars Liden and Kimin Lee and Jianfeng Gao and Luke Zettlemoyer and Dieter Fox and Minjoon Seo},
  booktitle     = {International Conference on Learning Representations (ICLR)},
  year          = {2025},
  eprint        = {2410.11758},
  archiveprefix = {arXiv}
}

@inproceedings{shentu2024lcb,
  title         = {From {LLMs} to Actions: Latent Codes as Bridges in Hierarchical Robot Control},
  author        = {Yide Shentu and Philipp Wu and Aravind Rajeswaran and Pieter Abbeel},
  booktitle     = {IEEE/RSJ International Conference on Intelligent Robots and Systems (IROS)},
  year          = {2024},
  eprint        = {2405.04798},
  archiveprefix = {arXiv}
}

@inproceedings{baker2022vpt,
  title         = {Video {PreTraining} ({VPT}): Learning to Act by Watching Unlabeled Online Videos},
  author        = {Bowen Baker and Ilge Akkaya and Peter Zhokhov and Joost Huizinga and Jie Tang and Adrien Ecoffet and Brandon Houghton and Raul Sampedro and Jeff Clune},
  booktitle     = {Advances in Neural Information Processing Systems (NeurIPS)},
  year          = {2022},
  eprint        = {2206.11795},
  archiveprefix = {arXiv}
}

@misc{unitree2024g1,
  title        = {Unitree {G1} Humanoid Robot},
  author       = {{Unitree Robotics}},
  year         = {2024},
  howpublished = {\url{https://www.unitree.com/g1}},
  note         = {Product page}
}

@inproceedings{mahmood2019amass,
  title         = {{AMASS}: Archive of Motion Capture as Surface Shapes},
  author        = {Mahmood, Naureen and Ghorbani, Nima and Troje, Nikolaus F. and Pons-Moll, Gerard and Black, Michael J.},
  booktitle     = {Proceedings of the IEEE/CVF International Conference on Computer Vision (ICCV)},
  year          = {2019},
  eprint        = {1904.03278},
  archiveprefix = {arXiv}
}

@article{mclean2025embody3d,
  title         = {Embody {3D}: A Large-scale Multimodal Motion and Behavior Dataset},
  author        = {McLean, Claire and Meendering, Makenzie and Swartz, Tristan and Gabbay, Orri and Olsen, Alexandra and Jacobs, Rachel and Rosen, Nicholas and de Bree, Philippe and Garcia, Tony and Merrill, Gadsden and Sandakly, Jake and Buffalini, Julia and Jain, Neham and Krenn, Steven and Kumar, Moneish and Markovic, Dejan and Ng, Evonne and Prada, Fabian and Saba, Andrew and Zhang, Siwei and Agrawal, Vasu and Godisart, Tim and Richard, Alexander and Zollhoefer, Michael},
  journal       = {arXiv preprint arXiv:2510.16258},
  year          = {2025},
  eprint        = {2510.16258},
  archiveprefix = {arXiv}
}

@inproceedings{ze2025twist,
  title         = {{TWIST}: Teleoperated Whole-Body Imitation System},
  author        = {Ze, Yanjie and Chen, Zixuan and Araujo, Joao Pedro and Cao, Zi-ang and Peng, Xue Bin and Wu, Jiajun and Liu, C. Karen},
  booktitle     = {Conference on Robot Learning (CoRL)},
  year          = {2025},
  eprint        = {2505.02833},
  archiveprefix = {arXiv}
}

@inproceedings{araujo2025gmr,
  title         = {Retargeting Matters: General Motion Retargeting for Humanoid Motion Tracking},
  author        = {Araujo, Joao Pedro and Ze, Yanjie and Xu, Pei and Wu, Jiajun and Liu, C. Karen},
  booktitle     = {IEEE International Conference on Robotics and Automation (ICRA)},
  year          = {2026},
  eprint        = {2510.02252},
  archiveprefix = {arXiv}
}

@article{lee2025phuma,
  title   = {Phuma: Physically-grounded humanoid locomotion dataset},
  author  = {Lee, Kyungmin and Kim, Sibeen and Park, Minho and Kim, Hyunseung and Hwang, Dongyoon and Lee, Hojoon and Choo, Jaegul},
  journal = {arXiv preprint arXiv:2510.26236},
  year    = {2025}
}

@inproceedings{kim2025pyroki,
  title         = {{PyRoki}: A Modular Toolkit for Robot Kinematic Optimization},
  author        = {Kim, Chung Min and Yi, Brent and Choi, Hongsuk and Ma, Yi and Goldberg, Ken and Kanazawa, Angjoo},
  booktitle     = {IEEE/RSJ International Conference on Intelligent Robots and Systems (IROS)},
  year          = {2025},
  eprint        = {2505.03728},
  archiveprefix = {arXiv}
}

@inproceedings{liu2024taco,
  title         = {{TACO}: Benchmarking Generalizable Bimanual Tool-{ACtion}-Object Understanding},
  author        = {Liu, Yun and Yang, Haolin and Si, Xu and Liu, Ling and Li, Zipeng and Zhang, Yuxiang and Liu, Yebin and Yi, Li},
  booktitle     = {IEEE/CVF Conference on Computer Vision and Pattern Recognition (CVPR)},
  year          = {2024},
  eprint        = {2401.08399},
  archiveprefix = {arXiv}
}

\clearpage
\appendix
\section{More Ablations}
\label{app:more-ablations}

\subsection{Rotation scale $\delta$} 
\label{app:more-abl-rotation-delta}
We vary the rotation scale $\delta$ for humanoid retargeting following the setup of Tab.~\ref{tab:exp:hh}. The choice $\delta=0.2$ reported in Sec.~\ref{sec:method:representation} of the main paper is empirical and is intended to keep the primal diffusion signal approximately normalized around mean $0$ and standard deviation $1$. Table~\ref{tab:more:delta} compares this choice with $\delta=0.6$ and $\delta=1.0$.

\begin{table}[htbp]
    \centering
    \caption{Rotation-scale ablation on humanoid retargeting.}
    \label{tab:more:delta}
    \begin{tabular}{lccc}
        \toprule
        Setting & SR-seq$\uparrow$ & gerr$\downarrow$ & lerr$\downarrow$ \\
        \midrule
        Paper ($\delta=0.2$) & 0.8690 & 0.5938 & 0.0420 \\
        $\delta=0.6$ & 0.8645 & 0.6440 & 0.0449 \\
        $\delta=1.0$ & 0.8030 & 0.7709 & 0.0504 \\
        \bottomrule
    \end{tabular}
\end{table}

\subsection{Semantic and geometric features}
\label{app:more-abl-feature}
We ablate $F_{\mathrm{sem}}$ and $F_{\mathrm{geo}}$ by setting the corresponding features to zero.

\paragraph{Language Table.}
Here, $F_{\mathrm{sem}}$ consists of three learnable embeddings for object, link-state, and link-action, while $F_{\mathrm{geo}}$ is obtained from a PointNet on the local point cloud of the object or link. Removing $F_{\mathrm{geo}}$ substantially reduces the success rate (Table~\ref{tab:more:lt-features}). The local colored point cloud grounds the language command in specific objects, whereas the semantic feature identifies the entity category rather than which object is which.

\begin{table}[htbp]
    \centering
    \caption{Feature ablations on Language Table. \texttt{noFsem} and \texttt{noFgeo} zero the respective features; \texttt{noBoth} zeros both.}
    \label{tab:more:lt-features}
    \begin{tabular}{lcccc}
        \toprule
         & Paper & \texttt{noFsem} & \texttt{noFgeo} & \texttt{noBoth} \\
        \midrule
        Avg Succ Rate$\uparrow$ & 0.8720 & 0.8560 & 0.1960 & 0.1680 \\
        \bottomrule
    \end{tabular}
\end{table}

\paragraph{TraceGen.}
Here, $F_{\mathrm{sem}}$ is the DINO feature queried at the pixel location. No $F_{\mathrm{geo}}$ is used because the prompt is a point. Removing DINO degrades performance on both EPIC and DROID (Table~\ref{tab:more:trace-features}), although the model can still predict future motion from the available geometric information. Semantic features encode “what” the entity is, improving the expressivity of the network.

\begin{table}[htbp]
    \centering
    \caption{Semantic-feature ablation on TraceGen.}
    \label{tab:more:trace-features}
    \small
    \setlength{\tabcolsep}{4pt}
    \begin{tabular}{lcccccc}
        \toprule
         & \multicolumn{3}{c}{EPIC} & \multicolumn{3}{c}{DROID} \\
        \cmidrule(lr){2-4}\cmidrule(lr){5-7}
        Setting & MAE$\downarrow$ & MSE$\downarrow$ & Endpoint$\downarrow$ & MAE$\downarrow$ & MSE$\downarrow$ & Endpoint$\downarrow$ \\
        \midrule
        Paper & 2.2564 & 0.3458 & 0.6146 & 1.1718 & 0.1665 & 0.2253 \\
        \texttt{noFsem} & 2.4032 & 0.3627 & 0.6148 & 1.2894 & 0.1876 & 0.2539 \\
        \bottomrule
    \end{tabular}
\end{table}

\subsection{Consistency with parametric models}
\label{app:more-abl-projection}
For evaluation, we convert predicted poses to each baseline's format: SMPL joint rotations computed from the $SE(3)$ poses, or robot joint angles obtained from relative axis-angles along the kinematic chain with one degree of freedom per joint. Evaluation then follows the corresponding baseline protocol. We measure the consistency between the raw $SE(3)$ predictions and their projected fits in two settings.

\paragraph{SMPL on the OMOMO test set.}
We compare raw $SE(3)$ joints and bones with the forward-kinematics skeleton of the projected SMPL parameters.

\begin{table}[htbp]
    \centering
    \caption{Self-consistency of SMPL predictions on the OMOMO test set.}
    \label{tab:more:smpl-consistency}
    \begin{tabular}{ccc}
        \toprule
        Bone length error & Bone direction error & Raw vs.\ projected MPJPE \\
        \midrule
        $7.21$\,mm & $2.99^\circ$ & $19.15$\,mm \\
        \bottomrule
    \end{tabular}
\end{table}

\paragraph{G1 humanoid test set.}
We measure the off-hinge-axis rotation and translation removed by projection, together with the difference between raw and forward-kinematics link positions.

\begin{table}[htbp]
    \centering
    \caption{Self-consistency of predictions on the G1 humanoid test set.}
    \label{tab:more:g1-consistency}
    \begin{tabular}{ccc}
        \toprule
        Off-hinge-axis rotation & Off-hinge translation & Raw vs.\ projected MPJPE \\
        \midrule
        $3.57^\circ$ & $13.37$\,mm & $31.38$\,mm \\
        \bottomrule
    \end{tabular}
\end{table}

\subsection{Data-generation cost}

For parametric and known embodiments, including SMPL bodies, MANO hands, and known robot URDFs, joint or link $SE(3)$ poses can be read directly from saved parameters or actions. 

To obtain pose labels for supervision from general RGB videos, we first preprocess the videos to obtain 3D tracks. Table~\ref{tab:more:data-cost} outlines the procedure and reports measurements for a video with $40$ frames, $32{,}768$ tracks, and resolution $378\times504$. The reported total includes the optional full-video object segmentation (VOS) stage, which has not yet been used in this paper.

After the tracking stage, we compute local frames from these tracks.
We first define a KNN topology using a topology-aware curve distance: the maximum distance between two trajectories over the times when they are co-visible. Local rigid motion is then obtained by solving a closed-form Procrustes problem.

\begin{table}[h]
    \centering
    \caption{RGB-video-to-3D-track processing cost.}
    \label{tab:more:data-cost}
    \small
    \setlength{\tabcolsep}{4pt}
    \begin{tabular}{@{}p{0.80\linewidth}p{0.20\linewidth}p{0.36\linewidth}@{}}
        \toprule
         Steps & Time, VRAM \\
        \midrule
        \textbf{Calibration}: resize and run GeoCalib on key frames to undistort intrinsics and estimate the gravity/vertical direction.
        & 4.3\,s / 5,856\,MB
        \\
        \midrule
        \textbf{Geometry}: run chunked slice-and-align DA3 (to keep VRAM bounded on long sequences) to estimate depth and camera parameters; also run DA3-metric on key frames to compute the metric scale.
        & 34.6\,s / 14,137\,MB
        \\
        \midrule
        \textbf{Tracking}: run 2D tracking using TAPNext and multiple rounds of motion-aware resampling (to keep VRAM bounded and focus more on moving areas), then lift the 2D tracks to 3D at visible time steps using the camera parameters and depth, with depth-boundary and anti-jitter filtering.
        & 64.6\,s / 8,151\,MB
        \\
        \midrule
        \textbf{[OPT] Full VOS (not used in the paper)}: our pipeline also runs DEVA in auto-mode to segment and track anything in the full video.
        & 27.2\,s / 14,079\,MB
        \\
        \midrule
        Total cost & 131\,s / 14.1\,GB \\
        \bottomrule
    \end{tabular}
\end{table}

\section{Technical appendices and supplementary material}
\label{app:overview}

\subsection{Related Work}
\label{sec:related}

\paragraph{Motion Generation and 4D World Modeling.}
\label{sec:related:motion}
Generative motion modeling has historically fragmented along actor classes. Body, human-human interaction, hand-object interaction, and scene dynamics each use their own state parameterization and task interface~\cite{tevet2022mdm,jiang2024motiongpt,barquero2024flowmdm,xu2023interdiff,li2024chois,diller2024cghoi}. A more recent line of work on unification scales LM backbones, million-scale corpora, compositional priors, and motion-understanding foundations on human motion alone~\cite{li2026llamomotion,zhang2024largemotionmodel,wang2025scalinglmm,baharani2025mofm,patel2025uniegomotion,xu2025multiperson,mu2025smp}, and three further papers consolidate human-side tasks under a single prior. MAGNet~\cite{maluleke2025magnet} models multi-agent body behavior with diffusion-forced canonical pairwise transforms, UMO~\cite{cong2026umo} casts object-aware human motion as a unified in-context problem, and FUSION~\cite{duran2026fusion} trains an unconditional SMPL-X body and hand prior. All of these inherit a body-parametric state and so remain actor-homogeneous within the human family. A second line writes motion into a shared observation space that crosses morphologies. MoMap~\cite{lei2025momaps} diffuses pixel-aligned dense 3D motion maps; PointWorld~\cite{huang2026pointworld} rolls point cloud and gripper flow as a forward simulator; TraceGen~\cite{lee2025tracegen} flow-matches camera-frame 3D traces conditioned on a single RGB-D start frame and language; GeneralFlow~\cite{yuan2024generalflow} predicts dense 3D flow from RGB-D; EC-Flow~\cite{chen2025ecflow} maps embodiment-centric pixel flow to end-effector $\mathrm{SE}(3)$ via inverse kinematics; ATM~\cite{wen2024atm} pretrains 2D point tracks as a representation for behavior cloning; TAPNext~\cite{zholus2025tapnext} casts point tracking itself as causal next-token prediction; and sparse-track or sparse-observation diffusion also extends to dexterous manipulation and SMPL completion~\cite{dex4d2026,castillo2023bodiffusion}. These methods share an observation space across actors, but the underlying tokens are translation-only points or dense flow that carry no per-entity identity and discard the rigid-body rotation that drives contact and articulation. A third line generates ambient 4D scene dynamics with explicit trajectories: Comp4D~\cite{xu2024comp4d} factorizes a 4D world into per-object trajectories plus per-object deformation, and TC4D~\cite{bahmani2024tc4d} conditions 4D generation on explicit $\mathrm{SE}(3)$ trajectories, but both target per-scene synthesis rather than a generative prior over heterogeneous moving entities. A piecewise-rigid lineage models scenes directly as per-part $\mathrm{SE}(3)$ transforms: SE3-Nets~\cite{byravan2017se3nets} and SE3-Pose-Nets~\cite{byravan2018se3posenets} warp tabletop point clouds through $K$ learned rigid motions conditioned on one robot's actions, OASIS~\cite{chen2026oasis} predicts the end-effector pose trajectory for manipulation, and SPOT~\cite{hsu2024spot} denoises source-target object pose trajectories from a few demonstrations, yet each remains a single-agent system over a fixed rigid-object set tied to one manipuland or one robot. The closest points of comparison to \ours are therefore MAGNet and UMO (unified conditioning, human-only), TraceGen and MoMap (trajectory-space world modeling without rotation or per-entity identity), and SPOT together with SE3-Nets ($\mathrm{SE}(3)$-trajectory denoising on a fixed manipuland set), each of which stays inside one of these regimes. \ours instead encode every moving unit, whether a robot link, hand, body, object, or scene primitive, as a sparse $\mathrm{SE}(3)$ token in a shared rot6d+translation space (Sec.~\ref{sec:method:representation}), so one denoiser is promptable across actor classes and queries arbitrary entities at arbitrary time steps, rather than tracking a fixed dense grid of points or a single morphology family.

\paragraph{World Models and Robot Policy.}
\label{sec:related:wm-vla}
A unified motion substrate must also support rollout and policy. Pixel-space world-action models keep the substrate in 2D video: DreamDojo~\cite{gao2026dreamdojo}, DreamZero~\cite{ye2026dreamzero}, and Cosmos~\cite{nvidia2025cosmos} synthesize frames or extract actions from pretrained video diffusion backbones, and World-VLA-Loop~\cite{liu2026worldvla} closes the action-perception loop on this pixel substrate with a vision-language-action policy. A related dense-observation model, DyWA~\cite{lyu2025dywa}, runs dynamics-adaptive world-action coupling over single-view point clouds for non-prehensile manipulation, swapping pixel tokens for a 3D point cloud but keeping the world model in a dense observation space. Latent-dynamics planners abstract observations into a learned latent representation: Hierarchical Humanoid WM~\cite{hansen2024hwm} and DreamerV3~\cite{hafner2023dreamerv3} couple latent rollouts with model-predictive control or value-based control, but each instance hard-codes the morphology or the target RL environment. Two recent approaches directly model heterogeneous forward dynamics: Cross-Embodiment WM~\cite{he2025crossembodimentwm} runs particle-graph rollouts and Geom4D-Robot~\cite{liu2025geom4drobot} jointly denoises RGB and pointmap, yet both recover $\mathrm{SE}(3)$ only post-hoc through an additional pose-extraction step rather than treating $\mathrm{SE}(3)$ as the modeling primitive; imagination-policy goal generators~\cite{huang2024imaginationpolicy} and long-horizon scene or traffic generators~\cite{zhan2026perpetualwonder,autoworld2026} likewise stay morphology- or domain-locked. Vision-language-action policies form a complementary set of output interfaces over various tokenizations of action~\cite{reed2022gato,black2024pi0,zhen2024threedvla}, and Moto~\cite{chen2024moto} bridges video pretraining to action through a latent motion-token language. Most closely related to our unification approach, UWM~\cite{zhu2025uwm} and UVA~\cite{li2025uva} jointly denoise video and action and pursue multi-axis flexibility through shared backbones, but both still tie the world-model side to pixel tokens. \ours close this gap by predicting $\mathrm{SE}(3)$ trajectories of every tracked entity under a per-token conditioning mask, so one flow-matching model with per-token noise serves as a policy, predictor, and world simulator across arm, humanoid, dexterous hand, human, and tabletop settings, with no morphology-specific head and no separate world and policy modules (Sec.~\ref{sec:method:inference}).

\paragraph{Sequence Modeling with Flexible Conditioning.}
\label{sec:related:df}
Diffusion forcing~\cite{chen2024diffusionforcing} assigns each token an independent noise level, so any subset of tokens can be revealed, in-filled, or rolled out from one trained network, and CausalFusion~\cite{deng2024causaldit} reconciles autoregressive and diffusion training under the same per-token schedule. This primitive has propagated along several tracks: streaming video adopts diffusion forcing for long-horizon rollouts and live generation~\cite{huang2025selfforcing,liu2025rollingforcing,huang2025memoryforcing}; motion and robot control adopt the same schedule on body, action, or interaction tokens~\cite{cai2026flooddiffusion,huang2024multimodaldf,hoeg2024streamingdp,forge2025difftree}; discrete and masked sequence models reach the same flexible-conditioning behavior through token-level masking schedules~\cite{arriola2025blockdiffusion,wang2025discretedf,chang2022maskgit}; a separate trajectory-level denoising line conditions whole state-action plans on goals or rewards~\cite{janner2022diffuser,chi2023diffusionpolicy}; and earlier flexible-conditioning recipes provide classifier-free guidance, masked image inpainting, and masked motion completion~\cite{ho2022cfg,lugmayr2022repaint,tessler2024maskedmimic}. Each of these instances, however, binds the per-token schedule to a single domain: video frames, one actor's skeletal tokens, one morphology's action horizon, or text. \ours apply the same per-token noise schedule to $\mathrm{SE}(3)$ trajectories of heterogeneous entities (Sec.~\ref{sec:method:training}) and extend the trajectory with a context token prefix on a separate timeline for non-causal anchors such as goal states or retargeting exemplars (Sec.~\ref{sec:method:representation}), so one trained network selects between motion synthesis, future prediction, policy rollout, and planning based solely on which tokens are noised at inference (Sec.~\ref{sec:method:inference}).

\subsection{Dense generation beyond the trained budget}
\label{app:dense-sampler}

\ours are trained on a bounded number of entities and time steps, yet many uses require more entities or time steps than fit in one forward pass (even though RoPE helps to generalize beyond the context window): long-horizon rollouts, dense scene flow, and multi-agent simulations. Splitting into independent chunks fails because two chunks denoised independently disagree at their interface. The \emph{anchored sliding-window} sampler in Alg.~\ref{alg:method:dense} instead exploits asynchronous-noise training to keep partially denoised tokens of one chunk as fixed anchors when denoising the next. With a context size $S$ and an anchor count $A<S$, the sampler descends a discrete schedule $\lambda_K>\cdots>\lambda_0=0$, so that every forward pass sees $S$ entities at consistent asynchronous levels, exactly the regime used in training with~\eqref{eq:method:loss}. The $A$ shared anchors stitch successive batches into a globally consistent rollout. Asynchronous training plus a non-grid token interface together unlock arbitrarily many entities and time steps at inference.

\begin{algorithm}[h]
\caption{Anchored sliding-window dense sampler.}
\label{alg:method:dense}
\begin{algorithmic}[1]
\Require $S,\,A\!<\!S,\,\lambda_K\!>\!\cdots\!>\!\lambda_0\!=\!0,\,\mathcal{E}$ at $\lambda_K$
\For{$k=K,\ldots,1$}
    \State $\mathcal{D}\gets\emptyset$
    \State pick $S$ at $\lambda_k$; advance to $\lambda_{k-1}$; add to $\mathcal{D}$
    \While{any entity at $\lambda_k$}
        \State $A$ anchors $\sim\mathcal{D}$; $S{-}A$ fresh at $\lambda_k$
        \State advance fresh to $\lambda_{k-1}$; add to $\mathcal{D}$
    \EndWhile
\EndFor
\State \Return $\mathcal{E}$ at $\lambda_0\!=\!0$
\end{algorithmic}
\end{algorithm}

\subsection{MPC details}
\label{app:exp:lt:mpc}

At each ACT replanning step, the MPC mode will (i) run one text-conditioned plain inference pass to obtain an imagined goal trace; (ii) build $K{=}4$ candidate action chunks of length~$10$ (one baseline plus three perturbed by Gaussian noise of $\sigma{=}0.02$\,m on the $xy$ components only, with $z$ and rotation frozen); (iii) run the diffusion agent for $12$ inner SGD iterations on the candidate $xy$ components with all $10$ DDIM denoising steps differentiable and with the language conditioning removed during the optimization rollouts, scoring each candidate by the mean $\ell_2$ distance of object positions at the end of the horizon from the imagined goal; and (iv) execute the best candidate for $\texttt{act\_horizon}{=}10$ environment steps before replanning. The environment, environment seed, $200$-step horizon, and $50$ rollouts per family match the plain policy protocol.

\subsection{Ablation configurations}
\label{app:exp:abl}

\paragraph{Language Table.} \emph{no AdaLN} replaces the per-DiT-block shift/scale/gate modulation with a single zero-init linear input fusion at the token-grid input, forcing conditioning to propagate through residuals only. \emph{no per-token noise} replaces the asynchronous per-token noise schedule of Sec.~\ref{sec:method:training} with a single shared diffusion timestep $t$ per sample, broadcast to all valid positions; at inference, known tokens are clamped via a noised q-sample with soft blending at strength $0.5$. 

\paragraph{TraceGen.} The reported model for TraceGen uses the `tpsqspt` layer configuration as in Fig.~\ref{fig:method:arch}; \emph{no register pool} collapses the middle three layers to plain DiT at the wide dimension, removing the register-compression path and yielding the `tpaaapt` layer configuration; \emph{plain DiT} collapses all seven layers to plain DiT, removing every architectural primitive and yielding the `aaaaaaa` layer configuration.

\subsection{More Details}

A typical \ourssing transformer architecture uses 8 heads with an embedding dimension of 512 and an FFN comprising a 2-layer MLP with hidden dimension 2048. The attention types are applied sequentially, one type per block. The model is trained on each application dataset using AdamW, a learning rate of 5e-5, a weight decay factor of 1e-3, and a decay step at 300K.
For each setting, we train one set of weights on a single A6000 48GB GPU using the largest batch size that fits in memory: Language Table 64; TraceGen 4/12 for P=400/128; OMOMO 16; humanoid 8.

\end{document}